\documentclass{article} 
\usepackage{iclr2026_conference,times}

\usepackage{amsmath,amsfonts,bm}

\def\eqref#1{equation~\ref{#1}}

\def\1{\bm{1}}

\DeclareMathAlphabet{\mathsfit}{\encodingdefault}{\sfdefault}{m}{sl}
\SetMathAlphabet{\mathsfit}{bold}{\encodingdefault}{\sfdefault}{bx}{n}

\newcommand{\KL}{D_{\mathrm{KL}}}

\usepackage{hyperref}
\usepackage{url}
\usepackage{algorithm}
\usepackage{algorithmic}
\usepackage{amsmath,amsthm,amssymb}
\usepackage{color}
\usepackage{xcolor}

\newcommand{\beq}{\begin{equation}}
\newcommand{\eeq}{\end{equation}}
\newcommand{\beqs}{\begin{eqnarray}}
\newcommand{\eeqs}{\end{eqnarray}}
\newcommand{\barr}{\begin{array}}
	\newcommand{\earr}{\end{array}}
\newcommand{\bali}{\begin{aligned}}
	\newcommand{\eali}{\end{aligned}}

\newcommand{\Dc}[0]{\ensuremath{\mathcal{D}} }

\newcommand{\Lc}[0]{\ensuremath{\mathcal{L}} }
\newcommand{\Mc}[0]{\ensuremath{\mathcal{M}} }

\newcommand{\Sc}[0]{\ensuremath{\mathcal{S}} }

\newcommand{\Xc}[0]{\ensuremath{\mathcal{X}} }
\newcommand{\Yc}[0]{\ensuremath{\mathcal{Y}} }

\newcommand{\Ebb}[0]{\ensuremath{\mathbb{E}} }

\newcommand{\Rbb}[0]{\ensuremath{\mathbb{R}} }

\newcommand{\ie}[0]{\emph{i.e., }}

\newcommand{\eg}[0]{\emph{e.g., }}
\newcommand{\cf}[0]{\emph{cf. }}

\newcommand{\Jmat}[0]{\ensuremath{{\bf J}} }

\newcommand{\Mmat}[0]{\ensuremath{{\bf M}} }

\newcommand{\Pmat}[0]{\ensuremath{{\bf P}} }

\newcommand{\Wmat}[0]{\ensuremath{{\bf W}} }

\newcommand{\hv}[0]{\ensuremath{\boldsymbol{h}} }

\newcommand{\qv}[0]{\ensuremath{\boldsymbol{q}} }
\newcommand{\wv}[0]{\ensuremath{\boldsymbol{w}} }
\newcommand{\xv}[0]{\ensuremath{\boldsymbol{x}} }

\newcommand{\zv}[0]{\ensuremath{\boldsymbol{z}} }

\newcommand{\thetav}[0]{\ensuremath{\boldsymbol{\theta}} }

\newcommand{\phiv}[0]{\ensuremath{\boldsymbol{\phi}} }

\newcommand{\omegav}[0]{\ensuremath{\boldsymbol{\omega}} }

\newcommand{\bb}[0]{\color{blue}}

\newcommand{\kk}[0]{\color{black}}

\newcommand{\microsize}{\fontsize{5}{6}\selectfont}

\usepackage{scalerel}
\newcommand\mathbox[1]{\mathord{\ThisStyle{%
			\fboxsep3\LMpt\relax\kern1\LMpt\fbox{$\SavedStyle#1$}\kern1\LMpt}}}
\newcommand{\tightbox}[1]{\setlength{\fboxsep}{1pt}\fbox{$#1$}}

\usepackage{multirow}
\usepackage{makecell}
\usepackage[table]{xcolor}
\usepackage{subcaption}

\title{Fine-Grained Class-Conditional Distribution Balancing for Debiased Learning}

\author{Miaoyun Zhao \thanks{\texttt{miaoyun9zhao}@gmail.com. Code: \url{https://github.com/MiaoyunZhao/FG_CCDB.}}\\
Key Laboratory of Social Computing and Cognitive Intelligence\\
Dalian University of Technology\\
Liaoning, China \\
\And
Qiang Zhang \thanks{Corresponding author, \texttt{zhangq}@dlut.edu.cn.}\\
Key Laboratory of Social Computing and Cognitive Intelligence\\
Dalian University of Technology\\
Liaoning, China \\
}

\iclrfinalcopy 
\begin{document}

\maketitle

\begin{abstract}
	Achieving group-robust generalization in the presence of spurious correlations remains a significant challenge, particularly when bias annotations are unavailable.
	Recent studies on Class-Conditional Distribution Balancing (CCDB) reveal that spurious correlations often stem from mismatches between the class-conditional and marginal distributions of bias attributes. They achieve promising results by addressing this issue through simple distribution matching in a bias-agnostic manner. 
	However, CCDB approximates each distribution using a single Gaussian, which is overly simplistic and rarely holds in real-world applications. 
	To address this limitation, we propose a novel 
	Multi-stage data-Selective reTraining strategy (MST),
	which describes each distribution in greater detail using the hard confusion matrix.
	Building on these finer descriptions, we propose a fine-grained variant of CCDB, termed FG-CCDB, which enhances distribution matching through more precise confusion-cell-wise reweighting. FG-CCDB learns sample weights from a global perspective, effectively mitigating spurious correlations without incurring substantial storage or computational overhead.
	Extensive experiments demonstrate that MST serves as a reliable proxy for ground-truth bias annotations and can be seamlessly integrated with bias-supervised methods.
	Moreover, when combined with FG-CCDB, our method performs on par with bias-supervised approaches on binary classification tasks and significantly outperforms them in highly biased multi-class and multi-shortcut scenarios.
\end{abstract}

\section{Introduction}
\label{sec:intro}

Neural networks trained with standard Empirical risk minimization (ERM) \cite{Vapnik1998Statistical} often suffer from spurious correlations: shortcuts that are predictive of the target class in the training data but irrelevant to the true underlying classification function \cite{labonte2023towards}. 
Samples exhibiting such spurious correlations typically dominate the training distribution and form the majority groups, while samples with different or conflicting correlations constitute the minority groups \cite{radford2021learning}. 
This imbalance across groups is also referred to as biased data,
which results in poor ERM performance on the minority ones, sometimes even no better than random guessing \cite{shah2020pitfalls}.
Spurious correlations are prevalent in many high-stakes applications, including toxic comments identification \cite{borkan2019nuanced}, medical diagnosis \cite{castro2020causality}, and autonomous driving \cite{Pourkeshavarz_2024_CVPR}, where both robustness and fairness are critical but overlooked by conventional methods.
Take the traffic sign classification task as a vivid example \cite{liuavoiding}, in which the training data exhibits a strong bias: 99\% of stop signs appear in red, whereas stop signs of other colors are rare and constitute a minority group \cite{beery2018recognition}. Consequently, the classifier relies on the red color as a shortcut for recognizing stop signs, ignoring the textual “stop” features. This leads to biased predictions and poor generalization when the color cue is absent or misleading. \cite{arjovsky2019invariant, geirhos2020shortcut,beery2018recognition}. 
These challenges underscore the urgent need to develop classification methods that remain reliable across diverse data subgroups, especially in the presence of spurious correlations. 


One of the most effective strategies for improving robustness against spurious correlations is to retrain models using group-balanced subsets derived from bias annotations \cite{kirichenko2023last}. However, given the massive scale of modern datasets, manually labeling bias attributes is often prohibitively expensive, which motivates the development of annotation-free alternatives. 
Recent studies have shown that models trained with naïve ERM tend to favor biased solutions, which generalize poorly to minority groups --- offering a ``free lunch'' for bias modeling 
\cite{pezeshki2024discovering,puli2023don}. Accordingly, various methods have been developed to identify misclassified samples as belonging to minority groups. These approaches either explicitly highlight such samples or implicitly simulate group-balancing during the debiasing process to enhance group robustness \cite{NEURIPS2023_265bee74,pezeshki2024discovering,li2023bias,liu2021just}. 
However, they often rely on empirically chosen hyperparameters to control the upweighting of minority groups, which can easily lead to overemphasis on these groups and, in turn, degrade performance on the majority ones. As a result, held-out annotations are often required for effective hyperparameter tuning.
Recent research on Class-conditional distribution balancing (CCDB) \cite{zhao2025class} reveals that spurious correlations arise from the mismatches between class-conditional and marginal distributions (usually caused by bias cues), and addresses it by reweighting samples to minimize the mutual information between bias cues and class labels without hyperparameter searching.
However, CCDB performs coarse distribution matching by treating each distribution as a single Gaussian, which rarely holds in real-world applications. In practice, instances within the same class often exhibit multi-modal distributions due to hidden bias cues. Thus, this coarse matching fails to capture intra-class variations, leaving residual spurious correlations unaddressed.

To resolve these limitations, we propose a fine-grained distribution matching technique based on CCDB, termed Fine-Grained Class-Conditional Distribution Balancing (FG-CCDB), which achieves stronger mitigation of spurious correlations without relying on bias annotations.
Our approach is developed from two key perspectives:
($i$) \textbf{Fine-grained distribution description.} 
Inspired by the “free lunch” phenomenon in ERM --- where models tend to overfit to spurious correlations --- we introduce 
a Multi-stage data-Selective reTraining strategy (MST) for bias characterization,
which capable of tackling multi-shortcuts by relate the hard confusion matrix to bias-aligning and conflicting partitions, and employing a multi-stage, data-selective retraining strategy to enhance the reliability of these partition assignments, which iteratively refines predictions from the overfitted model. This process yields a confusion matrix that approximates the ground-truth group partition when spurious correlations arise from a single shortcut.
($ii$) \textbf{Fine-grained distribution matching.} 
Building on the confusion matrix identified by MST, we extend CCDB into a fine-grained formulation, termed Fine-Grained Class-Conditional Distribution Balancing (FG-CCDB). It provides a discrete multi-modal approximation of both class-conditional and marginal distributions, enabling precise mode-wise alignment and thus more thorough mitigation of spurious correlations than the original CCDB.  
\textbf{The main contributions of this work are as follows:}
($i$) We propose an annotation-free bias exploration method with multi-stage refinement, based on model overfitting, which generalizes beyond singular shortcut and serves as a reliable alternative to human annotations.
($ii$) We introduce FG-CCDB, a lightweight and scalable debiasing method that enables fine-grained mode-wise reweighting and is well-suited for multi-class classification and multi-shortcut mitigation.
($iii$) Extensive experiments show that our method matches or surpasses bias-supervised baselines, achieving strong performance without requiring bias annotations.

%

\section{Related work}
\label{sec:relate}

The related work is structured around the two core aspects of our contribution.


\vspace{-0.1cm}

\subsection{Bias exploration}

\vspace{-0.1cm}

Primary approaches define bias as texture \cite{bahng2020learning}, background \cite{Venkataramani_2024_WACV}, or image style \cite{li2025invariant}—features presumed irrelevant to class labels. These methods often rely on tailored architectures or training schemes to detect specific bias cues \cite{hong2021unbiased}, but generalize poorly to unknown biases.
To overcome this, recent data-driven strategies interpret bias as group imbalance or latent substructures. Some methods, like JTT \cite{liu2021just}, LfF \cite{nam2020learning}, and RIDGE \cite{pezeshki2024discovering}, identify bias via consistently misclassified (hard) samples under ERM. 
Others rely on model disagreement, \eg DebiAN \cite{li2022discover} iteratively trains a bias ``discoverer'' alongside a main classifier, XRM \cite{pezeshki2024discovering} uses a pair of biased auxiliary models to generate pseudo group labels across the training set, DDB \cite{ciranni2025diffusing} utilizes a diffusion model to generate bias-aligned data, which amplifies the bias reliance.
Other methods, such as GEORGE \cite{sohoni2020no}, apply unsupervised feature clustering to decompose each class into latent subgroups. 
Few of these methods conduct a thorough evaluation on the quality of bias prediction.
Another trend leverages vision-language models (e.g., CLIP \cite{radford2021learning}) to infer explainable bias attributes \cite{jain2023distilling,kim2024discovering,wilesdiscovering}, though they are constrained by predefined vocabularies and may miss unexpected biases.



\vspace{-0.3cm}

\subsection{Bias mitigation}


\textbf{Bias annotation dependent}.
With the assistance of bias annotations, a variety of methods have been developed to mitigate spurious correlations.
GroupDRO \cite{sagawadistributionally} groups data based on class and bias annotations and optimizes for the worst-group performance.
DFR \cite{kirichenko2023last} improves robustness by retraining only the last layer using a small, balanced validation set.
MAPLE \cite{zhou2022model} uses a measure based on validation set with explicit bias annotations to reweight training samples.
LISA \cite{yao2022improving} utilizes data augmentation technique to encourage bias-invariant features.
Though effective, relying on costly bias annotations limits their scalability in real applications.

\textbf{Bias-conflicting samples dependent.}
To mitigate spurious correlations without manual annotation, recent studies often leverage disagreements among auxiliary models to identify bias-conflicting samples and focus learning on them.\cite{nam2020learning, liuavoiding, chu2021learning, liu2021just}.
To better identify bias-conflicting samples, SELF \cite{labonte2023towards} proposes to split the training data and applying early stopping for effective bias-conflicting detection.
uLA \cite{tsirigotis2023group} leverages pretrained self-supervised models to extract bias-relevant information.
DPR \cite{han2024mitigating} uses the Generalized cross-entropy loss \cite{nam2020learning} to amplify model bias. 
However, these methods rely on empirically tuned parameters—often requiring a split of annotated subsets—and their simple binary partitioning into bias-aligned and bias-conflicting samples is insufficient to fully capture the structure of bias, ultimately limiting generalization.

\textbf{Bias-agnostic.}
Beyond bias-aware techniques, several bias-agnostic approaches have emerged, motivated by diverse perspectives\cite{puli2023don,jain2024improving}.
MASKTUNE \cite{asgari2022masktune}, ExMap \cite{chakraborty2024exmap}, and DaC \cite{Noohdani_2024_CVPR} reduce reliance on spurious features by identifying bias-related regions via heatmaps, which restricts their applicability to the image domain.
Stable learning approaches\cite{zhang2021deep,yu2023stable} treat spurious correlations as effects of unknown confounders and attempt to mitigate them by decorrelating features, though this is difficult to achieve in practice.
GERNE \cite{asaad2025gradient} leverages the gradient differences between two batches to identify a debiasing direction, along which the model is optimized.
CCDB \cite{zhao2025class} seeks to mitigate spurious correlations by minimizing the mutual information between spurious features and class labels via distribution matching. Although effective, its coarse matching strategy limits generalization performance.

\section{Our method}
\label{sec:our}

Our work builds on the existing method CCDB, which attributes spurious correlations to distribution mismatches and addresses them through sample reweighting without requiring bias annotations. However, CCDB performs distribution matching in a relatively coarse manner by modeling each distribution as a single Gaussian.
To enable more accurate alignment--and thereby more effective spurious correlations elimination--we propose a fine-grained extension. Specifically, we introduce a multi-stage data-selective retraining strategy (MST) that characterizes bias structure via the hard confusion matrix, allowing for a discrete multi-modal description of each distribution.
Based on these multi-modal distributions, we develop Fine-Grained Class-Conditional Distribution Balancing (FG-CCDB), which performs alignment at the mode level. 


We consider the task of predicting a label $y\in\Yc,\Yc= \{1, \dots, C\}$ based on an input $\xv\in\Xc,\Xc\subset \mathbb{R}^d$. 
Following prior work, we define a \textit{shortcut} as an explainable attribute (\eg color, background) that is spuriously correlated with class labels and highly predictive,
and focus on a more general setting in which each data point $(\xv,y)$ may be associated with one or more shortcuts. 
Motivated by \cite{NEURIPS2023_b0d9ceb3}, we use an auxiliary biased model to predict the bias label $s\in\Sc$, which share the same label space as $y$, \ie $|\Sc|=|\Yc|$. 
Note that our goal is for $s$ to capture general and harmful bias information that humans may not preconceive \cite{li2022discover}, rather than only physically interpretable attributes. The value $s$ represents spurious signals that an ERM model prefers over core features and that consequently cause evaluation failures.
$s = i$ denotes all spurious cues that cause samples from other classes to be misclassified as class $i$. These cues may correspond to interpretable shortcuts, combinations of multiple shortcuts, or entangled, uninterpretable patterns.
By combining $s$ and $y$, we partition the dataset into modes $\Mc=\Sc\times\Yc$, which exactly corresponde to the hard confusion matrix. When the bias corresponds to a single shortcut, this reduces to conventional group partitions. To distinguish our data partitioning approach from traditional group-based methods, we refer to the partitions derived from the confusion matrix as \textbf{modes}. Accordingly, diagonal entries represent majority (bias-aligning) modes, and off-diagonal entries correspond to minority (bias-conflicting) modes.
With the confusion matrix, one can infer a discrete multi-modal approximation of both the class-conditional and marginal distributions over bias information.
The goal is twofold: ($i$) to train a biased model that can effectively explore the underlying bias cues; ($ii$) to train a debiased model that invariant to bias information and achieves uniform performance across all modes.



\subsection{Bias exploration through overfitting}
\label{sec:subsec:bias_explor}

\begin{figure*}[t]
	\vspace{-0.6 cm}
	\setlength{\abovecaptionskip}{2.0pt}
	\centering
	\includegraphics[width=1\columnwidth]{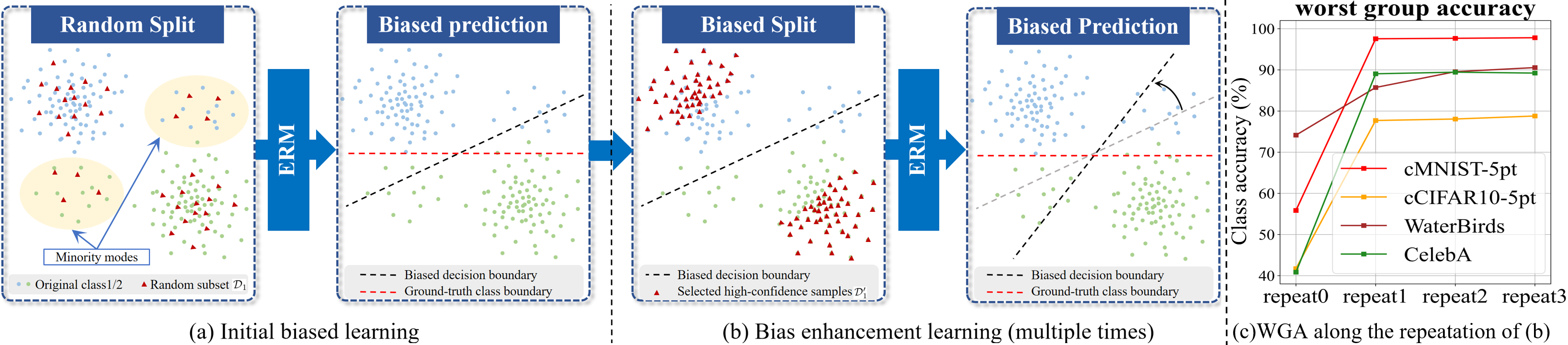}
	\vspace{-0.4 cm}
	\caption{The MST framework progresses from an initial partition with limited minority group coverage (a) to a more complete identification in stage (b). (c) WGA under different bias-capturing qualities.
	}\label{fig:framework_bof}
	\vspace{-0.2 cm}
\end{figure*}

In this section, we introduce the proposed multi-stage data-selective retraining (MST) technique and demonstrate its compatibility with existing bias-supervised methods.
It is well established that, in the presence of spurious correlations, ERM tends to overfit to majority groups in training data, leading to an over-reliance on bias cues and poor generalization to minority groups.  
Recent studies \cite{labonte2023towards,tsirigotis2023group} have made preliminary attempts to exploit this overfitting behavior to mitigate spurious correlations,  
revealing that the predictions of overfitted models are strongly aligned with bias cues.
Furthermore, \cite{lee2023revisiting,ciranni2025diffusing} find that removing bias-conflicting samples improves bias prediction and point out that, in principle, if all bias-conflicting samples were removed, one could train a bias-capturing model that provides ideal learning signals for debiasing.
Inspired by these insights, we propose a multi-stage framework for refined bias prediction, which further leverages model overfitting and serves as an approximate substitute for human annotations.
The overall framework consists of two basic stages (Figure \ref{fig:framework_bof}(a)(b)): initial bias learning and bias enhancement learning. 
The first stage extracts primary bias patterns, while the second amplifies them in the model’s predictions, yielding a reliable bias predictor.

\textbf{Initial bias learning.}
Given an accessible train dataset $\Dc:=\{(\xv_i,y_i)\}_{i=1}^N$ of $N$ samples and $C$ classes. 
Following prior works \cite{zhao2025class,labonte2023towards,pezeshki2024discovering}, we explore bias information by randomly splitting $\Dc$ into two subsets $\Dc_1$ and $\Dc_2$, where $\Dc_1$ contains a fraction $\gamma$ of the original data (Figure \ref{fig:framework_bof}(a), left). 
We then perform na\"ive ERM on $\Dc_1$ to train a biased model $f_{\thetav_1}$, which typically performs well on majority groups but poorly on minority groups. 
Unlike prior works\cite{labonte2023towards,pezeshki2024discovering}, which use $95\%$/$50\%$ of the data for biased training, our goal is to maximize the model's alignment with bias cues to better reveal underlying spurious correlations. As demonstrated in our experiments (Figure \ref{fig:correlate}(right)), a smaller $\gamma$ proves more effective for bias exploration, with $\gamma=10\%$ emerging as a sweet spot.
Since the data is randomly split, some samples from minority modes inevitably participate in training, which weakens the model’s tendency to align its predictions with bias cues (Figure \ref{fig:framework_bof}(a) right). To counteract this effect, we introduce a subsequent amplification stage.

\textbf{Bias enhancement learning.}
Amplifying bias in model predictions is non-trivial. Our key idea is to guide the bias prediction model to focus exclusively on majority modes, \ie to construct a training subset $\Dc_1$ that contains little to no samples from minority modes. This idea of removing bias-conflicting samples has been shown to effectively amplify bias in prior work \cite{lee2023revisiting,ciranni2025diffusing}. Such a setup forces the model to overfit to the majority modes and align more strongly with the corresponding bias cues, thereby behaving like a bias predictor and exhibiting near-zero generalization ability on minority modes. 
To achieve this, we introduce a data selection procedure based on the predictions of $f_{\thetav_1}$, forming an extremely biased subset $\Dc_1'$, on which a more biased model is trained. 
Specifically,
for each sample $(\xv_i,y_i)$ in $\Dc$, we infer the softmax output as $\hv_i=f_{\thetav_1}(\xv_i)$. 
Within each class, we select the top $\beta$ fraction of samples ($\beta\in[0,1]$) with the highest prediction confidence (measured by $\hv_i$), and aggregate them to form $\Dc_1'$. In our experiments, we find that $\beta=50\%$ offers a stable and reliable choice.
Since $f_{\thetav_1}$ is biased toward majority modes, the high-confidence samples are more likely to come from those modes. Consequently, $\Dc_1'$ filters out most minority mode instances and is thus more biased than $\mathcal{D}_1$. (Figure \ref{fig:framework_bof}(b) left).
We then train a new biased model $f_{\thetav_2}$ using na\"ive ERM on $\Dc_1'$.
The resulting model serves as the final bias predictor to produce bias labels for $\Dc$. 
Combined with the target class labels, the resulting hard confusion matrix yields estimated mode partitions over the space $|\Sc|\times|\Yc|$, which can serve as a proxy for group annotations in bias-supervised methods (Figure \ref{fig:framework_bof}(b) right).

Notably, the ``Bias enhancement learning'' stage can be repeated to further improve bias prediction accuracy. Only the biased model from the final repetition is used to generate bias labels. As shown in the experiments(Figure\ref{fig:framework_bof}(c)), a single iteration already achieves performance comparable to existing methods, while further iterations lead to gradually converging performance with diminishing gains.

\subsection{Fine-grained class-conditional distribution balancing }
\label{sec:subsec:FG_CCDB}

In this section, we present the Fine-grained Class-Conditional Distribution Balancing (FG-CCDB) approach.
With the hard confusion matrix obtained via MST, FG-CCDB improves both the quality of distribution matching and the efficiency of sample reweighting.


The original CCDB proposes to mitigate spurious correlations by directly minimizing the mutual information between bias cues and target classes, which is achieved by aligning each class-conditional distribution with the marginal distribution, while simultaneously balancing class proportions --- a generalization to traditional class balancing technique. Specifically, the objective is to minimize:

\vspace{-0.3cm}
\begin{equation}\label{eq:mutual}
	\Lc_\text{\omegav} 
	= I(\tightbox{\zv},y) - H(y) 
	=\Ebb_{p_{\omegav}(y)}\KL[p_{\omegav}(\tightbox{\zv}|y)\|p(\tightbox{\zv})] + \Ebb_{p_{\omegav}(y)}\log{p_{\omegav}(y)}
\end{equation}
\vspace{-0.4cm}

where $\tightbox{\zv}$ denotes the latent feature (with gradients detached) extracted by the biased model prior to the fully connected layer, which predominantly captures bias cues. $\omegav$ denotes the sample weights to be optimized, and $\KL[\cdot\|\cdot]$ refers to the Kullback–Leibler divergence \cite{kullback1951information}. 
Since the true distributions associated with $\zv$ and $y$ are unknown, CCDB approximates them using single Gaussian, which is insufficient for complex data with inherently multi-modal structures. Moreover, CCDB’s sample-level reweighting requires storing and processing feature representations for the entire dataset, incurring additional computational cost.

\begin{figure*}[t]
	\vspace{-0.8 cm}
	\setlength{\abovecaptionskip}{2.0pt}
	\centering
	\includegraphics[width=1\columnwidth]{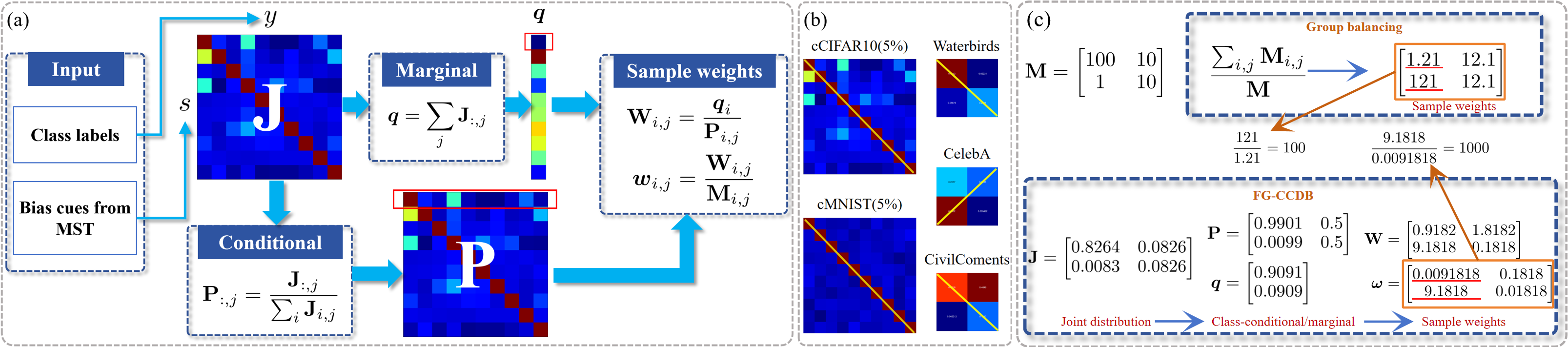}
	\vspace{-0.4 cm}
	\caption{(a) The framework of our FG-CCDB 
		(b) The joint distribution $\Jmat$ of bias and class labels estimated by our method.
		(c) Toy example to show how FG-CCDB differs from group balancing.
	}\label{fig:framework_fgccdb}
	\vspace{-0.4 cm}
\end{figure*}

Our work adopts the same objective as Equation \ref{eq:mutual}. To overcome the aforementioned limitations, we derive a discrete multi-modal approximation of both class-conditional and marginal distributions from the hard confusion matrix, which enables localized, mode-wise distribution matching and leads to more accurate and scalable reweighting.
As shown in Figure \ref{fig:framework_fgccdb} (a),
we represent the confusion matrix as $\Mmat\in\Rbb^{C\times C}$, where $\Mmat_{i,j}$ denotes the number of samples belonging to mode $(s,y)=(i,j)$, 
Thus, the joint distribution over $(\zv,y)$ is approximated with a discretized version over modes $(s,y)$, which is characterized by matrix $\Jmat\in\Rbb^{C\times C}$ with $\Jmat_{i,j}=\frac{\Mmat_{i,j}}{N}$ represents the probability of a sample belonging to mode $(s,y)=(i,j)$, $N$ is the total number of training samples. 
By design, we define a class-conditional distribution matrix $\Pmat\in\Rbb^{C\times C}$ such that the $j$-th column $\Pmat_{:,j}$ encodes $p(\zv|y=j)$, and a marginal distribution vector $\qv\in\Rbb^{C}$ that captures $p(\zv)$.
Both $\Pmat$ and $\qv$ are computed directly from $\Jmat$ as follows:
\vspace{-0.15cm}
\begin{equation}\label{eq:distr}
	p(\zv|y=j) \stackrel{\text{def}}{\approx} \Pmat_{:,j}=\frac{\Jmat_{:,j}}{\sum_{i}{ \Jmat_{i,j}}}, \:\:\:\:\:\:\:\:
	p(\zv) \stackrel{\text{def}}{\approx} \qv=\sum_{j}{ \Jmat_{:,j}}
\end{equation}
\vspace{-0.4cm}

Figure \ref{fig:framework_fgccdb} (b) shows the joint distribution matrix $\Jmat$ estimated by our MST across four datasets. Clear spurious correlations are observed, as evidenced by the strong diagonal elements (aligned along the yellow line), which indicate a high dependency between labels and bias cues.
To eliminate these spurious correlations and minimize equation\ref{eq:mutual}, we introduce mode-level weighting parameter $\Wmat\in\Rbb^{C\times C}$ to adjust each class-conditional distribution so that it aligns with the marginal distribution. 
A straightforward solution for $\Wmat$ is,

\vspace{-0.35cm}
\begin{equation}\label{eq:weightG}
	\Wmat_{i,j} =\frac{\qv_{i}}{\Pmat_{i,j}}\;, \qquad \text{for}\; i,j=1,\cdots,C
\end{equation}
\vspace{-0.3cm}

Note, equation \ref{eq:weightG} achieves exact distribution matching, \ie $\Wmat_{:,j}\odot\Pmat_{:,j}=\qv$, meaning that all class-conditional distributions are reweighted to align with the same marginal distribution $\qv$, where $\odot$ denotes the Hadamard product. 
For a given mode $(s,y)=(i,j)$, assuming uniform contribution from its samples, the corresponding sample weight is,
\vspace{-0.1cm}
\begin{equation}\label{eq:weightS}
	\wv_{i,j} =\frac{\Wmat_{i,j} }{\Mmat_{i,j}}\;, \qquad \text{for}\; i,j=1,\cdots,C
\end{equation}
\vspace{-0.3cm}

Note that beyond distribution matching, Equation \ref{eq:weightS} inherently solved the class imbalance issue: the mode with more samples gets smaller weights. As a result, FG-CCDB simultaneously minimizes both terms in Equation \ref{eq:mutual}. 
These weights are subsequently used during debiasing to reweight training samples according to their mode identities. 

It is worth noting that \textit{our distribution matching fundamentally differs from conventional group balancing} (see example in Figure\ref{fig:framework_fgccdb}(c)):
($i$) Unlike group balancing, which aims to reduce differences across all entries in the mode matrix, our method focuses solely on aligning the class-conditional distributions—i.e., reducing the variation among columns in $\Pmat$—while preserving intra-column imbalance. This allows for more flexible training by merely minimizing mutual information rather than enforcing strict equality.
($ii$) By minimizing the divergence between conditional and marginal distributions, our method and CCDB implicitly achieve ``covariate balance'' from the view of causal inference, specifically, by finding a reweighting that makes the confounder (bias) independent of the treatment (core feature), ultimately forcing the statistical model to rely solely on core features for inference \cite{neal2020introduction}.  
($iii$) Simple scale balancing between majority and minority modes is insufficient for generalization, as majority modes typically exhibit greater diversity. Our method applies a more aggressive reweighting strategy. For example, the ratio between the largest and smallest mode weights in FG-CCDB reaches 1000, compared to just 100 in conventional group balancing.
\textit{Compared to CCDB, our sample reweighting approach offers several key advantages}: 
($i$) It performs distribution matching across multiple localized regions defined by the confusion matrix, enabling more precise alignment and more thorough removal of spurious correlations; 
($ii$) The sample weights are computed efficiently in closed form, without requiring any iterative optimization; 
($iii$) Instead of assigning weights individually to each sample, FG-CCDB assigns a shared weight to samples within the same mode, resulting in negligible computational and memory overhead. 

After completing MST and FG-CCDB, we train a debiased model $f_{\phiv}$ by incorporating sample weights into the data sampling process using PyTorch's ``\textit{torch.utils.data.WeightedRandomSampler}'' following \cite{zhao2025class}. 
Unless otherwise specified, we refer to the entire procedure as FG-CCDB for brevity. 
A full algorithm of the proposed method is provided in Appendix \ref{appsec:algorithm}.


\section{Experimental results}
\label{sec:Experiments}


In this section, we demonstrate the effectiveness of our method from five perspectives:  
($i$) We conduct experiments on real-world binary classification benchmarks with either single or multiple shortcuts, such as Waterbirds \cite{zhou2022model}, CelebA \cite{zhou2022model}, CivilComments \cite{koh2021wilds}, and UrbanCars \cite{li2023whac} to validate the overall effectiveness of our method;
($ii$) We further evaluate our method on challenging multi-class datasets, including cMNIST \cite{li2022invariant} and cCIFAR10 \cite{hendrycks2018benchmarking} to assess its robustness under highly biased conditions;
($iii$) To evaluate the reliability of the bias cues explored by MST, we compare them with ground-truth bias annotations and analyze the effects of repeating the ``bias enhancement learning'' procedure; 
($iv$) we conduct an ablation study to demonstrate that each technical component (MST and FG-CCDB) makes a distinct and independent contribution to the final performance.
($v$) Finally, we analyze the effects of hyperparameters ($\gamma$ and $\beta$) on the performance of MST.

For all datasets, we adopt the same train-validation-test split following \cite{liu2021just,tsirigotis2023group} for fair comparison.
Results are averaged over 5 random seeds, and for each seed, the best-performing model (the one with the highest worst-class accuracy on the validation set) is selected \cite{tsirigotis2023group}. 
Unless otherwise stated, we repeat the ``bias enhancement learning'' process three times for FG-CCDB.
See the appendix for the detailed experimental setup.

\textbf{Compared methods.}
To demonstrate the superiority of our method in addressing spurious correlations and its potential to serve as an approximate substitute for bias-supervised methods, we compare it with both bias-supervised and bias-agnostic techniques. 
GroupDRO \cite{sagawadistributionally} and DFR \cite{kirichenko2023last} are fully bias-supervised during both training and validation, and serve as strong baselines. 
LfF \cite{nam2020learning}, JTT \cite{liu2021just}, LC \cite{liuavoiding}, DaC \cite{Noohdani_2024_CVPR},
and SELF \cite{labonte2023towards} rely on pseudo-bias supervision during training, but still require bias annotations during validation to achieve optimal performance.
In contrast, ERM, uLA \cite{tsirigotis2023group}, MASKTUNE \cite{asgari2022masktune}, XRM \cite{pezeshki2024discovering}, DebiAN, ExMap, DDB \cite{ciranni2025diffusing}, GERNE \cite{asaad2025gradient}, and CCDB \cite{zhao2025class}, similar to our method, are entirely bias-agnostic throughout both training and validation.

\begin{table*}[t]
	\vspace{-1.0 cm}
	\centering
	\caption{
		Classification performance on real-world datasets. We report the average test accuracy(\%) and std.dev. over 5 random seeds. Best bias-agnostic results in bold.
		\label{tab:waterbirds}
	}
	\vspace{-0.3 cm}
	\renewcommand{\arraystretch}{1.0}
	\resizebox{0.95\hsize}{!}{
		\setlength{\tabcolsep}{8pt}
		\begin{tabular}{c|cc|cc|cc|cc}
			\hline
			\multirow{2}{*}{Methods}& \multicolumn{2}{c|}{Bias label} & \multicolumn{2}{c|}{Waterbirds} & \multicolumn{2}{c|}{CelebA} & \multicolumn{2}{c}{CivilComments}\\
			\cline{2-9}
			&   Train & Val & i.i.d. & WGA & i.i.d. & WGA & i.i.d. & WGA \\
			\hline
			GroupDRO & Yes & Yes & 93.50 & 91.40 & 92.90 & 88.90  & 84.2 & 73.7 \\
			DFR & Yes & Yes & 94.20\scriptsize{$\pm$0.4} & 92.90\scriptsize{$\pm$0.2} & 91.30\scriptsize{$\pm$0.3} & 88.30\scriptsize{$\pm$1.1} & 87.2\scriptsize{$\pm$0.3} & 70.1\scriptsize{$\pm$0.8} \\
			LfF & No & Yes & 97.50 & 75.20  & 86.00 & 77.20 & 68.2 & 50.3\\
			JTT & No & Yes & 93.60 & 86.00  & 88.00 & 81.10 & 83.3 & 64.3\\
			LC & No & Yes & - & 90.50\scriptsize{$\pm$1.1}  & - & 88.10\scriptsize{$\pm$0.8} & - & 70.30\scriptsize{$\pm$1.2}\\
			SELF & No & Yes & - & 93.00\scriptsize{$\pm$0.3}  & - & 83.90\scriptsize{$\pm$0.9} & - & 79.10\scriptsize{$\pm$2.1}\\
			DaC & No & Yes & 95.3\scriptsize{$\pm$0.4} & 92.3\scriptsize{$\pm$0.4} & 91.4\scriptsize{$\pm$1.1} & 81.9\scriptsize{$\pm$0.7} & - & -\\ 
			\hline
			ERM & No & No & \textbf{97.30} & 72.60 &\textbf{ 95.60} & 47.20 & 81.6 & 66.7\\ 
			MASKTUNE & No & No & 93.00\scriptsize{$\pm$0.7} & 86.40\scriptsize{$\pm$1.9}& 91.30\scriptsize{$\pm$0.1} &  78.00\scriptsize{$\pm$1.2}  & - & -\\
			uLA & No & No & 91.50\scriptsize{$\pm$0.7} & 86.10\scriptsize{$\pm$1.5}& 93.90\scriptsize{$\pm$0.2} &  86.50\scriptsize{$\pm$3.7} & - & -\\ 
			XRM & No & No & 90.60 & 86.10 & 91.0 & 88.5 & 83.5 & 70.1\\ 
			DebiAN & No & No & 90.80 & 78.19 & 84.0 & 52.9 & - & - \\ 
			DDB & No & No & - & 90.34 & - & - & - & -\\ 
			GERNE & No & No & - & 89.88\scriptsize{$\pm$0.67} & - & 74.24\scriptsize{$\pm$2.51} & - & 63.10\scriptsize{$\pm$0.22}\\ 
			CCDB & No & No & 92.59\scriptsize{$\pm$0.10}  & 90.48\scriptsize{$\pm$0.28} & {90.08\scriptsize{$\pm$0.19}} & 85.27\scriptsize{$\pm$0.28} & 83.60\scriptsize{$\pm$0.21} & 75.00\scriptsize{$\pm$0.26} \\
			\textbf{FG-CCDB} & No & No & 92.50\scriptsize{$\pm$0.52}  & \textbf{90.56}\scriptsize{$\pm$0.24} & {89.71\scriptsize{$\pm$0.54}} & \textbf{89.22}\scriptsize{$\pm$0.19} 
			& \textbf{86.99}\scriptsize{$\pm$0.14} & \textbf{78.52}\scriptsize{$\pm$0.42}\\
			\hline
		\end{tabular}
	}
	\vspace{-0.15cm}
\end{table*}

\begin{table}
	\centering
	\begin{minipage}[t]{0.45\textwidth}
		\centering
		\caption{
			Results on UrbanCars. 
		}
		\label{tab:urbancars}
		\vspace{-0.3cm}
		\renewcommand{\arraystretch}{1.10}
		\resizebox{1\hsize}{!}{
			\setlength{\tabcolsep}{1.5pt}
			\begin{tabular}{c|cc|c|ccc}
				\hline
				\multirow{2}{*}{Methods}& \multicolumn{2}{c|}{Bias label} &\multirow{2}{*}{\makecell{I.D.\\ Acc}}& \multicolumn{3}{c}{Gap due to shortcuts($\uparrow$)} \\
				\cline{2-3}\cline{5-7}
				& Train & Val &  & BG & CoObj & BG+CoObj \\
				\hline
				GroupDRO & Yes & Yes & 91.6 & -10.9 & -3.6 & -16.4 \\
				\hline
				JTT & No & Yes & 95.9 & -8.1 & -13.3 & -40.1\\
				DaC & No & No & 98.17 & -3.78  & -9.78 & -58.58\\
				\hline
				ERM & No & No & 97.6 & -15.3 & -11.2 & -69.2 \\
				ExMap & No & No  & - & -5.9  & -9.9 & -30.7\\
				DebiAN & No & No & 98.0 & -14.9  & -10.5 & -69.0\\
				DDB & No & No & 86.39 & \textbf{-1.85} & \textbf{-0.52} & \textbf{-0.12} \\
				\textbf{FG-CCDB} & No & No  & 92.98  & \underline{-4.17} & \underline{-7.37} 
				& \underline{-4.9}\\
				\hline
			\end{tabular}
		}
	\end{minipage}
	\hfill
	\begin{minipage}[t]{0.54\textwidth}
		\centering
		\caption{
			Ablation study on four datasets. 
			\label{tab:ccdb_supervise}
		}
		\vspace{-0.3cm}
		\renewcommand{\arraystretch}{1.25}
		\resizebox{1\hsize}{!}{
			\setlength{\tabcolsep}{1pt}
			\begin{tabular}{l|cc|cc|c|c}
				\hline
				\multirow{2}{*}{Methods} & \multicolumn{2}{c|}{Waterbirds} & \multicolumn{2}{c|}{CelebA} & cMNIST & cCIFAR10\\
				\cline{2-7}
				& i.i.d. & WGA & i.i.d. & WGA  &i.i.d. &i.i.d.\\
				\hline
				GroupDRO & 93.50 & 91.40 & {92.90} & 88.90 & 84.20 & 57.32\\
				\rowcolor{gray!20}
				\makecell{GroupDRO\\-MST} & 90.82\microsize{$\pm$0.05} & 88.47\microsize{$\pm$0.35} & {88.69}\microsize{$\pm$0.15} & 85.21\microsize{$\pm$0.02} &84.07\microsize{$\pm$0.22}&55.73\microsize{$\pm$0.54}\\
				DFR & \textbf{94.20}\microsize{$\pm$0.4} & \textbf{92.90}\microsize{$\pm$0.2} & 91.30\microsize{$\pm$0.3} & 88.30\microsize{$\pm$1.1} &-&-\\
				\rowcolor{gray!20}
				DFR-MST & 92.53\microsize{$\pm$0.50} & 91.49\microsize{$\pm$0.72} & 88.80\microsize{$\pm$0.20} & 85.87\microsize{$\pm$0.29} &-&-\\
				\hline
				FG-CCDB & 92.50\microsize{$\pm$0.52}  & 90.56\microsize{$\pm$0.24} & {89.71\microsize{$\pm$0.54}} & \textbf{89.22}\microsize{$\pm$0.19} &98.21\microsize{$\pm$0.02}&78.06\microsize{$\pm$0.30}\\
				\rowcolor{gray!20}
				\makecell{FG-CCDB\\-sup} & 91.54\microsize{$\pm$0.11}  & {91.76}\microsize{$\pm$0.13} & \textbf{93.14}\microsize{$\pm$0.16} & 89.09\microsize{$\pm$0.12} &\textbf{98.26}\microsize{$\pm$0.21}&\textbf{78.53}\microsize{$\pm$0.37}\\
				\hline
			\end{tabular}
		}
	\end{minipage}
	\vspace{-0.4cm}
\end{table}

\subsection{Binary classification with a single or multiple shortcuts}

The results on real-world binary classification with a single shortcut are shown in Table\ref{tab:waterbirds}. Although i.i.d. performance reflects overall accuracy, it can mask disparities across groups. In contrast, worst-group accuracy (WGA) directly measures robustness by focusing on the most challenging subpopulations.  
With bias annotations available during both training and validation, GroupDRO and DFR demonstrate strong generalization performance on the worst group, serving as a challenging upper bound.
In contrast, methods that only use bias annotations during validation show a bit inferior performance.
The situation becomes more challenging when access to bias annotations is not permitted. In this case, existing bias-agnostic methods consistently fall short of the supervised upper bound on at least one of the datasets. 
Remarkably, SELF, CCDB and our method surpass the supervised upper bound on CivilComments by a large margin. This is because they apply stronger upweighting to the minority groups/modes.
Among all compared methods, including those with full supervision, our method consistently achieves the best or competitive WGA across all three datasets, highlighting its effectiveness in eliminating the need for human annotations.

Table\ref{tab:urbancars} presents the results on UrbanCars with multiple shortcuts: background (BG) and co-occurring object (CoObj). 
The in-distribution accuracy(I.D. Acc) and gap-related metrics are adopted from \cite{li2023whac}(See appendix for details). The BG/CoObj/BG+CoObj Gap is the drop in accuracy between mean and cases when only the BG/CoObj/BG+CoObj is uncommon. A smaller drop indicates better generalization.
On average, BG+CoObj is the most challenging one and most compared methods suffer a significant drop on it. 
GroupDRO can mitigate multiple shortcuts; however, they require access to labels of both shortcuts. 
Although DDB shows the smallest overall drops across all bias-conflicting scenarios, its base I.D. Acc is the lowest among all compared methods.
Overall, our method consistently achieves the best balance between high I.D. Acc and small drops compared to other bias-agnostic methods (particularly on the challenging BG+CoObj generalization). It performs comparably to, or better than, methods that rely on bias annotations.
These results confirm that our approach provides a general framework for handling multi-shortcut scenarios.
Please refer to Appendix \ref{secapp:urbancars} for more details.

\subsection{Multi-class classification under extreme spurious correlations}

In this section, we use the synthetic datasets cMNIST and cCIFAR10 to evaluate the effectiveness of our method under challenging multi-class settings with extreme spurious correlations.
For each, we vary the ratio of bias-conflicting samples in the training set to control the strength of spurious correlations and evaluate performance on a completely unbiased test set.
Following \cite{tsirigotis2023group}, the bias-conflicting ratios are set to $\{0.5\%,1\%,2\%,5\%\}$ for both datasets, where $0.5\%$ indicates an extremely biased scenario.
The generalization accuracies are reported in Table \ref{tab:CMNIST}. 
We observe that:
($i$) On both datasets, our method consistently achieves the best performance. In particular, it outperforms the second-best method by a large margin on cMNIST; 
($ii$) On cCIFAR10, the improvements become more pronounced as the bias-conflicting ratio increases (i.e., at $2\%$ and $5\%$). 

\begin{table*}[t]
	\vspace{-0.5 cm}
	\centering
	\caption{
		Results on cMNIST and cCIFAR10 with various bias-conflicting ratios in the training set. The test accuracy(\%) is averaged over 5 random seeds.
		The best results are indicated in bold.
		\label{tab:CMNIST}
	}
	\vspace{-0.3 cm}
	\renewcommand{\arraystretch}{1.1}
	\resizebox{0.99\hsize}{!}{
		\setlength{\tabcolsep}{3pt}
		\begin{tabular}{c|cc|cccc|cccc}
			\hline
			\multirow{2}{*}{Methods}& \multicolumn{2}{c|}{Bias label} & \multicolumn{4}{c|}{cMNIST} & \multicolumn{4}{c}{cCIFAR10}\\
			\cline{2-11}
			&   Train & Val & $0.5\%$ & $1\%$ & $2\%$ & $5\%$ & $0.5\%$ & $1\%$ & $2\%$ & $5\%$\\
			\hline
			GroupDRO & Yes & Yes & 63.12 & 68.78 & 76.30& 84.20 & 
			33.44 & 38.30 & 45.81 & 57.32  \\
			LfF & No & Yes & 52.50\scriptsize{$\pm$2.43} & 61.89\scriptsize{$\pm$4.97} & 71.03\scriptsize{$\pm$2.44} & 80.57\scriptsize{$\pm$3.84} & 
			28.57\scriptsize{$\pm$1.30} & 33.07\scriptsize{$\pm$0.77} & 39.91\scriptsize{$\pm$0.30} & 50.27\scriptsize{$\pm$1.56}  \\
			LC & No & Yes & 71.25\scriptsize{$\pm$3.17} & 82.25\scriptsize{$\pm$2.11} & 86.21\scriptsize{$\pm$1.02} & 91.16\scriptsize{$\pm$0.97} &
			34.56\scriptsize{$\pm$0.69} & 37.34\scriptsize{$\pm$0.69} & 47.81\scriptsize{$\pm$2.00} & 54.55\scriptsize{$\pm$1.26}  \\
			DaC & No & Yes & 53.24 & 75.02 & 87.60 & 94.70 &
			21.01 & 28.01 & 36.56 & 51.06  \\
			\hline
			ERM & No & No & 35.19\scriptsize{$\pm$3.49} & 52.09\scriptsize{$\pm$2.88} & 65.86\scriptsize{$\pm$3.59} & 82.17\scriptsize{$\pm$0.74}& 23.08\scriptsize{$\pm$1.25} & 25.82\scriptsize{$\pm$0.33} & 30.06\scriptsize{$\pm$0.71} & 39.42\scriptsize{$\pm$0.64}  \\
			uLA & No & No & 75.13\scriptsize{$\pm$0.78} & 81.80\scriptsize{$\pm$1.41} & 84.79\scriptsize{$\pm$1.10} & 92.79\scriptsize{$\pm$0.85} & 
			34.39\scriptsize{$\pm$1.14} & 62.49\scriptsize{$\pm$0.74} & 63.88\scriptsize{$\pm$1.07} & 74.49\scriptsize{$\pm$0.58} \\
			GERNE & No & No & 77.25\scriptsize{$\pm$0.17} & 83.98\scriptsize{$\pm$0.26} & 87.41\scriptsize{$\pm$0.31} & 90.98\scriptsize{$\pm$0.13} & 
			39.90\scriptsize{$\pm$0.48} & 45.60\scriptsize{$\pm$0.23} & 50.19\scriptsize{$\pm$0.18} & 56.53\scriptsize{$\pm$0.32} \\
			CCDB & No & No & 83.20\scriptsize{$\pm$2.17} & 87.95\scriptsize{$\pm$1.59} & 91.02\scriptsize{$\pm$0.28} & 96.37\scriptsize{$\pm$0.25} & 
			55.07\scriptsize{$\pm$0.85} & 63.28\scriptsize{$\pm$0.46} & 67.78\scriptsize{$\pm$0.78} & 74.64\scriptsize{$\pm$0.34} \\
			\textbf{FG-CCDB} & No & No & \textbf{89.02}\scriptsize{$\pm$0.45} & \textbf{94.93}\scriptsize{$\pm$0.17} & \textbf{96.18}\scriptsize{$\pm$0.19} & \textbf{98.21}\scriptsize{$\pm$0.02} & 
			\textbf{55.28}\scriptsize{$\pm$0.54} & \textbf{64.66}\scriptsize{$\pm$0.48} & \textbf{71.69}\scriptsize{$\pm$0.31} & \textbf{78.06}\scriptsize{$\pm$0.30} \\
			\hline
		\end{tabular}
	}
	\vspace{-0.3 cm}
\end{table*}

A comparison of the results in Table \ref{tab:waterbirds} and Table \ref{tab:CMNIST} reveals a phenomenon similar to that reported in \cite{zhao2025class}: bias-supervised methods tend to perform well on basic binary classification tasks, whereas bias-agnostic methods are relatively more effective in complex multi-class classification scenarios. 
In contrast, CCDB demonstrates strong performance across both scenarios. With fine-grained distribution matching, FG-CCDB further boosts performance over CCDB by a significant margin, highlighting the effectiveness of more thorough spurious correlations elimination.

\begin{figure}[b]
	\vspace{-0.4 cm}
	\setlength{\abovecaptionskip}{2.0pt}
	\centering
	\includegraphics[width=0.999\columnwidth]{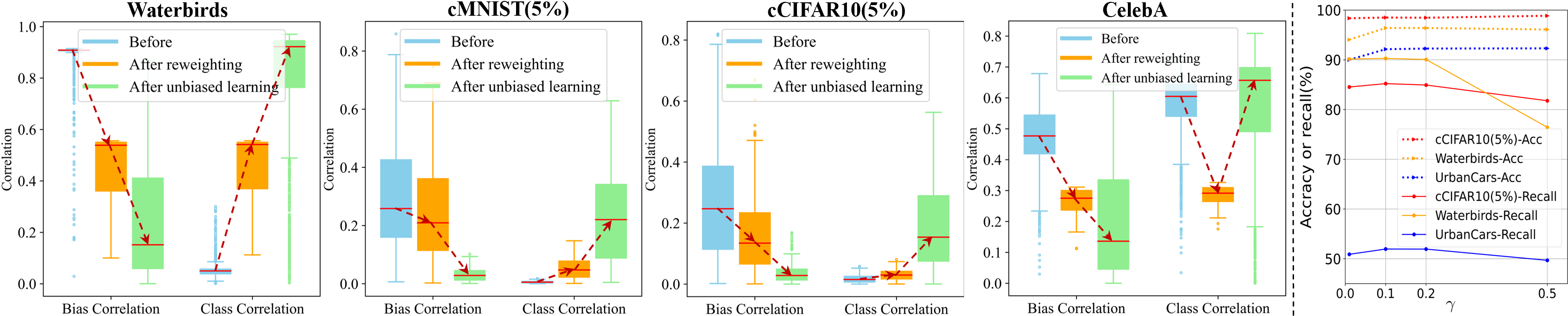}
	\vspace{-0.2 cm}
	\caption{\textbf{Left:} effect of FG-CCDB sample reweighting in reshaping the data distribution and mitigating spurious correlations. 
		\textbf{Right:} mode prediction accuracy and Smallest-mode recall along $\gamma$.
	}\label{fig:correlate}
	\vspace{-0.3 cm}
\end{figure}

To demonstrate the effectiveness of FG-CCDB in mitigating spurious correlations, we analyze how sample reweighting influences the correlation between feature dimensions and class/bias information, as shown in Figure \ref{fig:correlate}(left).
We compute the correlation of each feature dimension with class/bias information, and visualize their distributions using box plots \cite{zhao2025class}. Before sample reweighting, strong spurious correlations in the training data lead the biased model $f_{\thetav_2}$ to rely heavily on bias-related features, with most dimensions exhibiting high correlation with bias and low correlation with class. After applying FG-CCDB weights on features from $f_{\thetav_2}$, the correlation with bias drops significantly, while the correlation with class increases.
Moreover, after debiasing training on the reweighted data, this shift toward class-relevant features is further amplified, confirming that FG-CCDB effectively reduces the model's reliance on spurious features.

\subsection{The quality of bias exploration}


\begin{figure*}[t]
	\vspace{-0.9 cm}
	\setlength{\abovecaptionskip}{2.0pt}
	\centering
	\includegraphics[width=1\columnwidth]{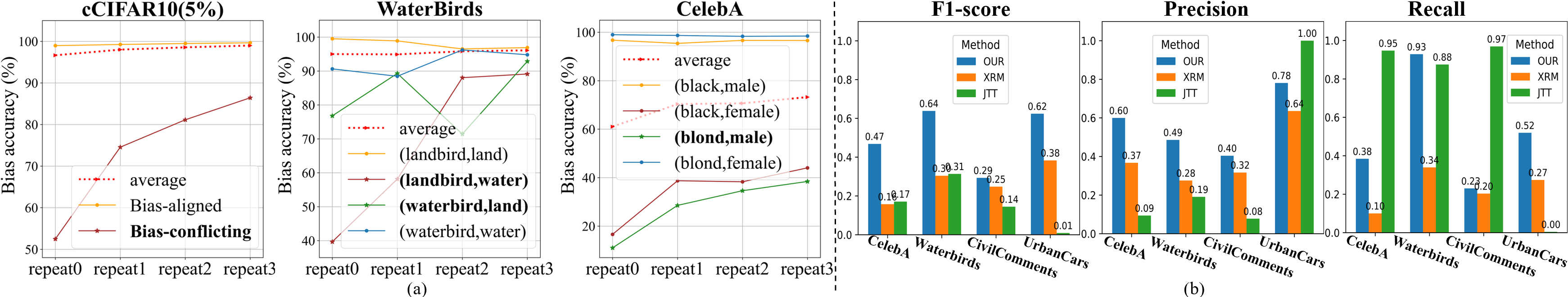}
	\vspace{-0.3 cm}
	\caption{(a) The mode prediction accuracy along the repeating of the ``bias enhancement learning'' procedure;
		(b) Smallest-mode F1-score, precision, and recall compare with existing methods.
	}\label{fig:repeat_bias}
	\vspace{-0.2 cm}
\end{figure*}

In this section, we evaluate the effectiveness of MST by measuring its mode-prediction F1-score, precision, and recall against the ground-truth annotations.
Results regarding the smallest-mode are shown in Figure\ref{fig:repeat_bias}(b). 
Since our method progressively filters out bias-conflicting samples, it retains far fewer such samples than XRM and JTT, achieving the highest F1-score across the four datasets. This confirms the principle that removing bias-conflicting samples improves bias prediction.
JTT misidentifies a large number of majority samples as belonging to the smallest-mode(low precision). In contrast, XRM tends to misidentify minority-modes samples as majority modes(low recall). 
With multi-stage refinement, our method achieving the best overall performance.
As discussed in Section \ref{sec:our}, repeating the ``bias enhancement learning'' process can further improve both bias prediction accuracy and consequently mode prediction accuracy. To validate this claim, we conduct experiments with different numbers of repetitions. The mode prediction performance across varying repetition counts are shown in Figure \ref{fig:repeat_bias}(a). 
The dashed lines represent the standard accuracy across all modes, while the solid lines show the recall for each individual mode. We observe that repetition has a particularly strong effect on minority groups (highlighted in bold), as evidenced by the significant improvement in their recall with more repetitions. Please refer to Appendix Figure \ref{figapp:repeating} for convergence results with additional repetitions.

Figure \ref{fig:framework_bof}(c) shows the final WGA for classification as repetition increases. Notably, performance improves substantially after the first repetition and then plateaus, especially on cMNIST and CelebA, suggesting that a single repetition is often sufficient to achieve satisfactory performance.



\subsection{Ablation study}


Our method comprises two core technical modules: MST and FG-CCDB, which together demonstrate superior performance.
In this section, we integrate these modules with existing methods and observe the resulting performance improvements to verify the effectiveness and versatility of our approach, as detailed below.

($i$) To assess the effectiveness of MST, we replace ground-truth annotations in bias-supervised methods, \ie GroupDRO and DFR, with bias predictions generated by MST. This results in their unsupervised counterparts, denoted as GroupDRO-MST and DFR-MST, respectively.
The results are reported at the top of Table \ref{tab:ccdb_supervise}.
Remarkably, the generalization performance of these unsupervised variants is comparable to their supervised versions using human annotations.
Although the bias predictions are not perfect, they are sufficiently accurate to identify most minority modes, confirming the effectiveness of our MST as an approximate substitute for human bias annotations.


($ii$) To evaluate the effectiveness of FG-CCDB independently of MST, we replace the predicted bias with human annotations, resulting in a supervised version, FG-CCDB-sup. 
The results are reported at the bottom of Table \ref{tab:ccdb_supervise}.    
When bias annotations are available, FG-CCDB-sup further boosts  performance, achieving results comparable to existing supervised methods on Waterbirds, and outperforming them on the others, especially in multi-class settings. \textit{This justified our statement on a more aggressive reweighting and indicates that FG-CCDB is a more effective strategy than na\"ive group balancing for handling spurious correlations}.
Moreover, the performance gap between FG-CCDB-sup and the original FG-CCDB is marginal, further confirming the effectiveness of our method in reducing reliance on human bias annotations.

\subsection{Hyperparameter Analysis}

In this section, we evaluate the effect of the hyperparameters $\gamma$ for ``Initial Bias Learning'' and $\beta$ for selecting top high-confidence samples on MST's final performance. The results are presented in Figure~\ref{fig:correlate}(right) and Table \ref{tab:top_confidence}. 

The hyperparameter $\gamma$ controls the proportion of samples selected for training the initial bias model. Intuitively, a smaller $\gamma$ leads to stronger overfitting to bias cues and thus greater reliance on them. As expected, the results in Figure \ref{fig:correlate}(right) show that when $\gamma \le 0.2$, both prediction accuracy and smallest-mode recall remain high. However, as $\gamma$ increases to $0.5$, the performance drops significantly. We find that $\gamma = 0.1$ serves as a sweet spot,  while also saving computation compared to $\gamma=0.2$. 

F1-score with $\beta \in \{30\%,50\%,70\%\}$ are reported in Table \ref{tab:top_confidence}. 
The hyperparameter $\beta$ controls the proportion of top high-confidence samples selected to filter out bias-conflicting samples and amplify the model's bias. Intuitively, this value relates to the smallest bias-aligned ratio across classes, as shown in the first row of Table~\ref{tab:top_confidence}. Except for CelebA, whose ratio is slightly above $50\%$, all other datasets have ratios exceeding $90\%$. Accordingly, $\beta=50\%$ serves as a reasonable middle-ground choice. For CelebA, which has a relatively low bias-aligned ratio, $\beta = 50\%$ achieves the best performance; whereas for datasets with ratios exceeding $90\%$, both $\beta = 50\%$ and $\beta = 70\%$ yield high F1-scores, with $\beta = 70\%$ performing the best. Intuitively, when bias annotations are unavailable, selecting the top $50\%$ high-confidence samples is likely to capture the bias-aligned subset while excluding bias-conflicting samples. 

\begin{table*}[t]
	\centering
	\caption{
		The F1-score of the smallest-mode prediction under different top high-confidence ratio $\beta$.
		\label{tab:top_confidence}
	}
	\vspace{-0.3 cm}
	\renewcommand{\arraystretch}{1.1}
	\resizebox{0.65\hsize}{!}{
		\setlength{\tabcolsep}{6pt}
		\begin{tabular}{c|c|c|c|c}
			\hline
			& cCIFAR10($5\%$) & Waterbirds & CelebA & UrbanCars \\
			\hline
			Bias-align ratio & 95.00\% & 94.97\% & 51.72\% & 90.25\% \\
			\hline 
			$\beta=30\%$ & 0.65 & 0.53 & 0.32 & 0.47 \\
			$\beta=50\%$ & 0.72 & 0.64 & \textbf{0.47} & 0.62 \\
			$\beta=70\%$ & \textbf{0.79} & \textbf{0.67} & 0.40 & \textbf{0.64}\\
			\hline
		\end{tabular}
	}
	\vspace{-0.4 cm}
\end{table*}

\section{Conclusions}
\label{sec:conclusions}

In this paper, we address the challenge of robust group generalization under spurious correlations without requiring bias annotations.
Following the distribution matching paradigm, we propose a method that integrates a reliable bias prediction module with fine-grained class-conditional distribution matching. Our approach demonstrates strong performance on real-world datasets with single or multiple shortcuts, as well as highly biased multi-class datasets, often matching or outperforming methods that rely on human-provided group annotations.
By leveraging the model’s overfitting behavior, our method offers a novel alternative to traditional group balancing strategies and effectively reduces reliance on manual supervision. 
However, its effectiveness may be limited in scenarios where the overfitting signal fails to capture bias cues—for example, in CelebA, which has only one minority group, or in CivilComments, where majority groups dominate one class while minority groups appear in another. These settings present different spurious correlation patterns that weaken the overfitting signal used for bias prediction.
Addressing this limitation remains an important direction for future research.

\subsubsection*{Acknowledgments}
This work was supported by the National Key R\&D Program of China under Grant No. 2024YFA1012700.
We thank Chenrong Li and Ruolan Liu for their assistance with figure preparation.

\bibliography{ReferencesCong}
\bibliographystyle{iclr2026_conference}

\newpage

\appendix{Appendix for \\Fine-Grained Class-Conditional Distribution Balancing for Debiased Learning}

\section{The algorithm of our proposed method}
\label{appsec:algorithm}

The complete procedure of our proposed FG-CCDB is summarized in Algorithm \ref{alg:fg-ccdb}. 
It consists of three stages: bias exploration, sample weight inference, and unbiased classifier training. 

\begin{algorithm}[h]
	\caption{
		Fine-grained class-conditional distribution balancing (FG-CCDB)
	}\label{alg:fg-ccdb}
	\begin{algorithmic}
		\STATE {\bfseries Input:} Randomly initialized network $f_{\thetav_1}$ and $f_{\thetav_2}$ for bias prediction, $f_{\phiv}$ for unbiased classification;
		training set $\Dc$, validation set $\Dc_v$.
		\STATE {\bfseries Output:} unbiased classifier $f_{\phiv}$.
		\STATE  $\#$\bb \textit{Stage1: bias exploration via multi-stage data-selective retraining}\kk
		\STATE \textbf{1:} Randomly sample a subset $\Dc_1$ from $\Dc$ with proportion $\gamma$ ($\gamma=10\%$). \\
		\STATE \textbf{2:} Train $f_{\thetav_1}$ on $\Dc_1$ using ERM.
		\STATE \textbf{3:} Select the top $\beta$ ($\beta=50\%$) most biased samples from $\Dc$ to form an extremely biased subset $\Dc_1'$.
		\STATE \textbf{4:} Train $f_{\thetav_2}$ on $\Dc_1'$ using ERM.
		\STATE \textbf{5:} use $f_{\thetav_2}$ to infer bias labels for all samples in $\Dc$ , and modeling joint distribution via hard confusion matrix.
		\STATE  $\#$ \bb\textit{Stage2: Sample weight inference}\kk
		\STATE \textbf{6:} Infer class-conditional and marginal distribution over the bias cues using equation\ref{eq:distr}.
		\STATE \textbf{7:} Compute sample weights using Equation\ref{eq:weightG}, and \ref{eq:weightS} from the main manuscript. 
		\STATE  $\#$ \bb\textit{Stage 3: Unbiased classifier training}\kk
		\STATE \textbf{8:} Train classifier $f_{\phiv}$ on reweighted samples using standard ERM. \\
		\STATE \textbf{9:} Select the best-performing $f_{\phiv}$ based on the highest worst-class accuracy on the validation set $\Dc_v$.
	\end{algorithmic}
\end{algorithm}

\section{Experimental setup}

\textbf{Datasets.}
The experiments are conducted on five benchmark datasets known to exhibit spurious correlations. 
Waterbirds, CelebA, CivilComments, and UrbanCars are real-world datasets in which each class is spuriously correlated with background, gender, certain demographic identities, or a combination of multiple shortcuts respectively. 
cMNIST and cCIFAR10 are synthetic ten-way classification tasks, where each class is spuriously linked to a specific color or noise pattern. 
For all datasets, we adopt the same train-validation-test split following \cite{liu2021just,tsirigotis2023group} for fair comparison.

\textbf{Training setup.}
For fair comparison, we adopt model architectures following \cite{tsirigotis2023group, labonte2023towards}: a 3-hidden layer MLP for cMNIST, ResNet18 \cite{he2016deep} For cCIFAR10, ResNet50 \cite{he2016deep} for Waterbirds and CelebA, and BERT \cite{devlin2019bert} for CivilComments.
ResNet18 and ResNet50 are pretrained on ImageNet-1K, and BERT is pretrained on Book Corpus and English Wikipedia.
No data augmentation is applied to cMNIST and CivilComments, while simple augmentations (random cropping and horizontal flipping) are used to the remaining datasets, following \cite{ahuja2021invariance}.
This ensures that the improvements we observed are attributed to the proposed methodology, rather than to data augmentations that could potentially nullify the bias attribute.
For our method, both the initial bias learning and the bias enhancement learning span 20 epochs, and the final unbiased learning involves 5000 iterations across all datasets. 
Results are averaged over 5 random seeds, and for each seed, the best-performing model (the one with the highest worst-class accuracy on the validation set \cite{tsirigotis2023group}) is selected. 
Unless otherwise stated, we repeat the ``bias enhancement learning'' process three times for FG-CCDB.

\textbf{On Hyperparameters.}
All experiments were conducted on a single NVIDIA A40 GPU.
The hyperparameters and optimization settings for the MST and FG-CCDB modules on each dataset are summarized in Table \ref{tab:hyper}.
Both modules share the same batch size, scheduler, optimizer, and optimizer hyperparameters.  
For cMNIST and CivilComments, no data augmentation is applied to either module, while for the remaining datasets, simple data augmentations (\ie ResizedCrop and HorizontalFlip) are applied only for the FG-CCDB module.
All stages in MST are trained with the same \textit{Epoch} number.
In contrast to CCDB, our MST framework consists of at least two stages: the first stage provides an initial bias prediction, which is further refined by the subsequent stages. 
The experimental results in the main manuscript (see Figure\ref{fig:correlate}(right)) show that selecting the parameter $\gamma$ within the range of $1\% \le \gamma \le 20\%$ has a negligible impact on the final performance.
Accordingly, we set $\gamma = 10\%$ across all datasets to ensure strong performance while maintaining low computational cost.

\begin{table*}[ht]
	\centering
	\caption{
		The optimization setup for our FG-CCDB. 
	}
	\label{tab:hyper}
	\renewcommand{\arraystretch}{1.5}
	\resizebox{1.0\hsize}{!}{
		\setlength{\tabcolsep}{2pt}
		\begin{tabular}{c|cccccccc}
			\hline \hline
			Dataset & Optimizer & Scheduler & LR & Batch size & Weight decay & $\{\text{Epoch,Iter}\}$ & $\gamma$  & Augmentation  \\
			\hline
			cMNIST &Adam & None & $1\times10^{-2}$&256&$1\times10^{-4}$&$\{20,5000\}$& 0.1 &None \\
			cCIFAR10 &Adam & None &$1\times10^{-5}$&256&$1\times10^{-4}$&$\{20,5000\}$& 0.1 & ResizedCrop, HorizontalFlip \\
			Waterbirds &Adam & None &$1\times10^{-5}$&256&$1\times10^{-4}$&$\{20,5000\}$& 0.1 & ResizedCrop, HorizontalFlip \\
			CelebA &Adam & None &$1\times10^{-5}$&256&$1\times10^{-4}$&$\{20,5000\}$& 0.1 & ResizedCrop, HorizontalFlip \\
			CivilComments &AdamW & Linear &$1\times10^{-5}$&16&$1\times10^{-4}$&$\{20,5000\}$& 0.1 & None \\
			\hline \hline
		\end{tabular}
	}
\end{table*}

\begin{figure*}[h]
	\setlength{\abovecaptionskip}{2.0pt}
	\centering
	\includegraphics[width=1\columnwidth]{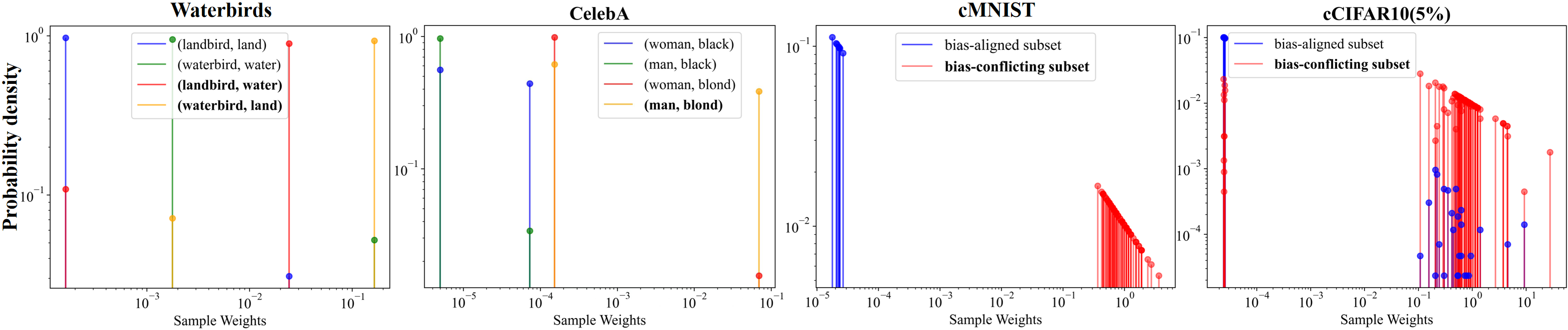}
	\caption{The distribution of the sample weights assigned by FG-CCDB within each mode on four datasets.
	}\label{fig:samp_weight}
\end{figure*}

\section{The sample weights inferred by our FG-CCDB}

To assess whether our distribution-matching approach, FG-CCDB, effectively distinguishes minority modes from majority ones and assigns appropriate sample weights in the singular shortcut case, we analyze the distribution of inferred sample weights across different modes. The results on four datasets are summarized in Figure \ref{fig:samp_weight}, with the minority modes highlighted in bold.
As expected, samples from the majority modes are assigned low weights, typically concentrated below 0.01, while samples from minority modes receive significantly higher weights, clustered around 1. These results demonstrate that FG-CCDB successfully differentiates between majority and minority modes, and up-weights the latter in a balanced manner, aligning both class-conditional and marginal distributions.


\section{Additional Experimental Results and Details for UrbanCars}
\label{secapp:urbancars}

\begin{figure*}[h]
	\setlength{\abovecaptionskip}{2.0pt}
	\centering
	\includegraphics[width=0.6\columnwidth]{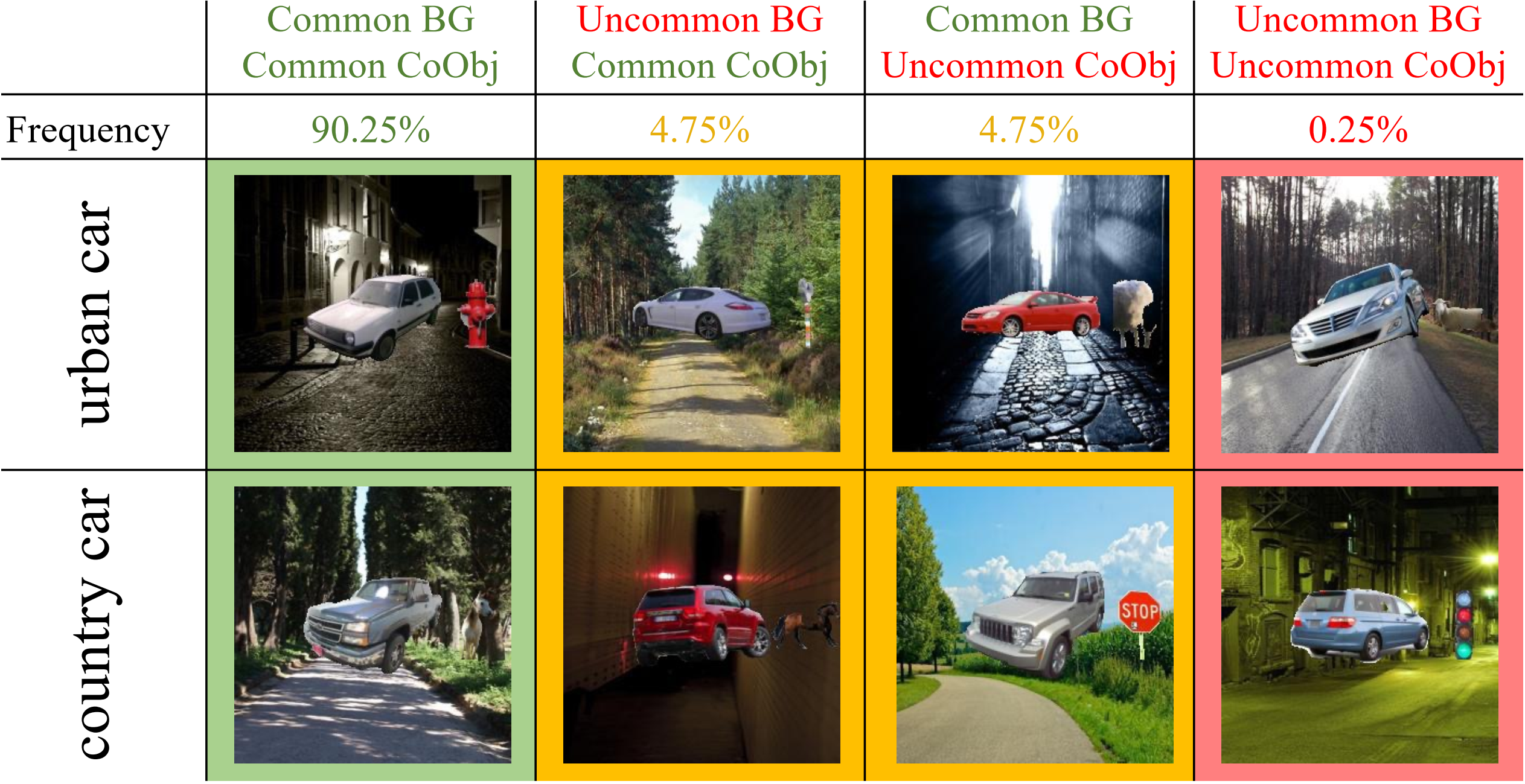}
	\caption{Unbalanced groups in the UrbanCars training set based on two shortcuts: background and co-occurring object (the figure is adopted from \cite{li2023whac})
	}\label{fig:Urbancars}
\end{figure*}

For UrbanCars, the class label corresponds to the car type (country or urban), while the spurious attributes consist of two shortcuts: the background (BG) and the co-occurring object (CoObj), both of which are also labeled as country or urban.
The ground-truth group partition of the training data is shown in Figure\ref{fig:Urbancars}. The majority groups contain urban car images combined with urban backgrounds (\eg alleys) and urban co-occurring objects (\eg fire plugs), and vice versa for country car images. The remaining combinations constitute the minority groups.
As shown in \cite{li2023whac}, mitigating spurious correlations in datasets with multiple shortcuts presents a Whac-A-Mole dilemma: mitigating one shortcut often amplifies the model's reliance on the others.

\textbf{Evaluation Metrics for the UrbanCars Dataset.}
Compared to datasets with a single shortcut, four new metrics are proposed for multi-shortcut scenarios to better evaluate performance across different shortcut combinations.

($i$) In-Distribution Accuracy (I.D. Acc):
This metric computes the weighted average of per-group accuracies, where the weights are proportional to each group's frequency in the training set (\ie its correlation strength, as shown in Figure\ref{fig:Urbancars}). Following the ``average accuracy'' definition in \cite{sagawadistributionally}, it reflects model performance when no group shift occurs.

($ii$) BG Gap:
The drop in accuracy from the I.D. Acc to the accuracy on groups where the background (BG) is uncommon but the co-occurring object (CoObj) remains common (\cf the first yellow column in Figure\ref{fig:Urbancars}).

($iii$) CoObj Gap:
The drop in accuracy from the I.D. Acc to the accuracy on groups where the CoObj is uncommon but the BG remains common (\cf the second yellow column in Figure\ref{fig:Urbancars}).

($iv$) BG+CoObj Gap:
The drop in accuracy from the I.D. Acc to the accuracy on groups where both BG and CoObj are uncommon (\cf the red column in Figure\ref{fig:Urbancars}).

BG Gap and CoObj Gap measure the model's robustness to distribution shifts caused by each individual shortcut.
BG+CoObj Gap evaluates robustness in the most challenging scenario, where both shortcuts are absent.

Following \cite{chakraborty2024exmap}, two variants of UrbanCars are constructed:
($i$) UrbanCars (BG), where only the background object serves as the spurious attribute;
($ii$) UrbanCars (CoObj), where only the co-occurring object serves as the spurious attribute.

We compare the worst-group accuracy (WGA) on these two variants plus the original one, as shown in Table\ref{aptab:urbancars}.
Our method achieves significantly higher WGA than ExMap on both variants, further confirming our claim that FG-CCDB captures bias information through mode partitioning in a more general manner. This makes it applicable to both singular and multiple shortcut scenarios.

\begin{table*}[t]
	\centering
	\caption{
		Classification performance on multi-shortcuts UrbanCars. In addition to our worst-group accuracy, the measurements following \cite{li2023whac} are also provided.
		\label{aptab:urbancars}
	}
	\renewcommand{\arraystretch}{1.1}
	\resizebox{0.95\hsize}{!}{
		\setlength{\tabcolsep}{6pt}
		\begin{tabular}{c|c|c|ccc||cc|cc|cc}
			\hline
			\multirow{2}{*}{Methods}& \multirow{2}{*}{Given Condition} &\multirow{2}{*}{I.D. Acc}& \multicolumn{3}{c||}{Gap due to shortcut} &\multicolumn{2}{|c}{Urbancar(BG)} & \multicolumn{2}{|c}{Urbancar(CoObj)} &\multicolumn{2}{|c}{Urbancar} \\
			\cline{4-12}
			&    &  & BG & CoObj & BG+CoObj & Mean & WGA & Mean & WGA & Mean & WGA \\
			\hline
			LfF & Yes  & 97.2 &-11.6 & -18.4 & -63.2 & - & - & - & -& - & - \\
			JTT & Yes  & 95.9 & -8.1 & -13.3 & -40.1 & - & - & - & -& - & - \\
			DebiAN & No  & 98.0 & -14.9  & -10.5 & -69.0 & - & - & - & -& - & - \\
			ExMap & No  & - & -5.9  & -9.9 & -30.7 & 93.2 & 71.4 & 93.2 & 79.2 & - & -\\
			\textbf{FG-CCDB} & None & 92.98  & \textbf{-4.17} & \textbf{-7.37} 
			& \textbf{-4.9} &91.04\scriptsize{$\pm$0.04}& 87.84\scriptsize{$\pm$0.2.12} & 93.08\scriptsize{$\pm$0.14} & 90.24\scriptsize{$\pm$0.28} & 88.56\scriptsize{$\pm$0.30} & 81.28\scriptsize{$\pm$3.9}\\
			\hline
		\end{tabular}
	}
	\vspace{-0.4 cm}
\end{table*}

\section{Additional Discussions}

\textbf{R1W1: How iterative bias amplification improves minority-mode recall}
 
In addition to our experimental results, the validation of MST is supported by the following research findings:
($i$) Easy-to-learn property of bias attributes \cite{nam2020learning}. ERM tend to overfit spurious correlations only when they are "easier" to learn than the desired core features.
This property has been successfully exploited in many debiasing methods \cite{nam2020learning,pezeshki2024discovering,labonte2023towards,zhao2025class,lee2023revisiting} to detect and highlight underrepresented bias-conflicting samples. 
Thus, the initial step of MST is well motivated.
($ii$) Removing bias-conflicting samples improves bias prediction. Prior works \cite{lee2023revisiting,ciranni2025diffusing} show that even a small number of bias-conflicting samples can severely degrade the estimation of bias-aligned vs. bias-conflicting partitions. In principle, if all bias-conflicting samples were removed, one could train a bias-capturing model that provides ideal learning signals for debiasing. These methods obtain a bias-amplified model either by explicitly removing bias-conflicting samples or by generating only bias-aligning samples. 
Our MST shares the same core insight but adopts a different mechanism: we use a multi-stage bias amplification process that progressively filters out bias-conflicting samples by selecting those with the highest confidence. 
($iii$) Bias-aligned samples tend to have higher confidence. As revealed in \cite{lee2023revisiting}, bias attributes are easier to learn than intrinsic attributes; thus, ERM model assigns higher predicted probabilities to bias-aligned samples. This phenomenon has also been effectively used in works on GCE \cite{zhang2018generalized}.
Therefore, selecting top-confidence samples at each stage in MST is an effective strategy for filtering out bias-conflicting samples.



\textbf{R1Q1: Why fix the top-$50\%$ high-confidence samples per-class for bias enhancement?}

We denote by $\beta$ the ratio used to select the top high-confidence samples for brevity. 
Our choice of $\beta=50\%$ is based on a practical and widely observed property of spurious-correlation datasets. In typical settings, within each class, the bias-aligned partition is larger than the bias-conflicting partition; otherwise, spurious correlations would not arise, as pointed out in \cite{ciranni2025diffusing}. This implies that the bias-aligned partition occupies more than $50\%$ of the samples in that class. 
Table \ref{tabapp:top_confidence} summarizes the smallest bias-aligned ratio across classes for each dataset. Except for CelebA, which has a value only slightly above $50\%$, the other datasets have ratios exceeding $90\%$. 
Therefore, when bias annotations are unavailable, selecting the top $50\%$ high-confidence samples is highly likely to capture the bias-aligned partition while excluding bias-conflicting samples. We emphasize that this is an empirical principle rather than a strict theoretical guarantee. However, it is consistently supported by prior works on spurious correlations and by our empirical results. 

To further address potential concerns regarding $\beta=50\%$, we conduct experiments with alternative proportions ($30\%$ and $70\%$) and an adaptive version based on class-wise confidence distributions (assigning higher $\beta$ to classes with higher average confidence). The F1-scores are shown in Table \ref{tabapp:top_confidence}. Clearly, $\beta=50\%$ represents a reasonable middle-ground option.
For CelebA, which has a low bias-aligned ratio, $\beta=50\%$ performs best, whereas for datasets with bias-aligned ratios above $90\%$, $\beta=70\%$ yields the best performance.
The adaptive strategy is primarily effective when the data exhibits noticeable class imbalance. We consider further exploration of this approach as promising future work.
\begin{table*}[t]
	\centering
	\caption{
		The F1-score of the smallest-mode prediction under different top high-confidence ratio $\beta$.
		\label{tabapp:top_confidence}
	}
	\renewcommand{\arraystretch}{1.1}
	\resizebox{0.65\hsize}{!}{
		\setlength{\tabcolsep}{6pt}
		\begin{tabular}{c|c|c|c|c}
			\hline
			& cCIFAR10($5\%$) & Waterbirds & CelebA & UrbanCars \\
			\hline
			Bias-align ratio & 95.00\% & 94.97\% & 51.72\% & 90.25\% \\
			\hline 
			$\beta=30\%$ & 0.65 & 0.53 & 0.32 & 0.47 \\
			$\beta=50\%$ & 0.72 & 0.64 & \textbf{0.47} & 0.62 \\
			$\beta=70\%$ & \textbf{0.79} & \textbf{0.67} & 0.40 & \textbf{0.64}\\
			\hline 
			Adaptive & 0.76 & 0.69 & 0.43 & 0.66 \\
			\hline
		\end{tabular}
	}
	\vspace{-0.4 cm}
\end{table*}

\textbf{R2W1: How iterative bias amplification improves minority-mode recall}

Please refer to R1W1.

\textbf{R2W2: comparison with recent label-free debiasing methods}

We incorporate comparisons with recent label-free debiasing methods: DDB \cite{ciranni2025diffusing}, DaC \cite{Noohdani_2024_CVPR}, and GERNE \cite{asaad2025gradient}.
DDB utilizes a diffusion model to generate bias-aligned data, which amplifies the bias reliance of the bias model and provides useful information for the debiasing process.
DaC identifies the causal components of images using class activation maps from models trained with ERM. It then intervenes on the images by combining these components and retrains the model on the augmented data.
Both DDB and DaC are specifically designed for image data.
GERNE assumes that the difference between the gradients of two batches captures a debiasing direction and optimizes the model along this direction.
The results are summarized in Table1, Table2 and Table4.  
Although DaC uses bias annotations during validation, its performance on CelebA remains significantly lower than ours. Our method demonstrates substantial advantages over GERNE and DDB across CelebA, CivilComments, and the multi-shortcut UrbanCars dataset. Notably, on UrbanCars, while DDB exhibits the smallest overall drops across different bias-conflicting scenarios, its base I.D. accuracy is the lowest among all compared methods.

\textbf{R2W3: The performance on multi-bias scenarios}
The experiments on multi-bias (multi-shortcut) scenarios may have been overlooked. We conducted experiments on the UrbanCars dataset, which contains multiple shortcuts (i.e., background and co-occurring objects). The corresponding results and discussion can be found in Section 4.1 and Table 2. 

Overall, our method consistently achieves the best balance between high I.D. accuracy and minimal drops compared to other bias-agnostic methods, particularly on the challenging BG+CoObj generalization. It performs comparably to --- or better than --- methods that rely on bias annotations.
These results confirm that our approach provides a general framework for handling multi-shortcut scenarios.

\textbf{R2W4: whether FG-CCDB can compensate for imperfect bias predictions}

We have shown the performance of FG-CCDB under different mode partition qualities in Figure \ref{fig:framework_bof}(c), which may have been overlooked.
By observing Figure 4(a), we find that repetition has a particularly strong effect on minority groups: performance increases significantly after the first repetition and then gradually converges.
Accordingly, in Figure \ref{fig:framework_bof}(c), the WGA obtained by subsequent FG-CCDB shows a similar trend: it jumps from a relatively low accuracy after the first repetition and then gradually converges to a stable value.
We conclude that:
($i$) When MST provides poor mode partitioning ("repeat0"), the errors are significant, and FG-CCDB is affected by these errors, resulting in relatively low WGA.
($ii$) When MST provides acceptable mode partitioning (with a repetition count of 1 or higher), the WGA of FG-CCDB increases and shows only marginal improvement with further repetitions, even though the mode partition quality continues to improve. This indicates that FG-CCDB can compensate for imperfect mode partitions once the partition quality is sufficiently high.

\textbf{R2Q1: how well the MST matches human labels? performance comparison results with the latest methods}

Please refer to R3W3 and R3W4 for a quantitative evaluation of MST's performance.
Please refer to R2W2 for a comparison with the latest label-free and generative model-based methods.

\textbf{R3W1: Definition of ‘mode’ and whether major biases are captured by MST}

We define the "mode" $(s,y)$ as a black-box concept because our goal is for $s$ to capture general and harmful bias information that humans may not preconceive \cite{li2022discover}, rather than only physically interpretable attributes. The value $s$ represents spurious signals that an ERM model prefers over core features and that consequently cause evaluation failures. We do not aim to model spurious attributes are not preferred by ERM and therefore do not lead to generalization errors. In this sense, model mistakes serve as indicators of harmful spurious correlations.
Regarding the type of bias we focus on, we clarify that \textbf{our model is unlikely to fail to capture such harmful bias cues}. The reasons are as follows.

First, extensive prior works \cite{nam2020learning,pezeshki2024discovering,labonte2023towards,zhao2025class,lee2023revisiting} operate under the widely accepted assumption that naive ERM tends to misclassify or produce low-confidence predictions on bias-conflicting samples. These studies demonstrate that ERM naturally learns spurious correlations, providing reliable learning signals for debiasing.

Second, for stronger theoretical grounding, we connect our idea to the Equal Opportunity Fairness (EOF) criterion \cite{li2022discover,hardt2016equality} and show that our method is equivalent to find the bias cues that cause a classifier’s predictions to strongly violate this fairness criterion, as detailed below.

Formally, a classifier $f$ satisfies EOF criterion if:
\begin{equation}
	\text{Pr}\{\hat{y}=k|s=0,y=k\} = \text{Pr}\{\hat{y}=k|s=1,y=k\}
\end{equation}
where the LHS and RHS are the true positive rates (TPR) for negative ($s=0$) and positive ($s=1$) groups in target class $k\in \{1...K\}$. As noted in \cite{li2022discover}, 
a significant TPR discrepancy between groups indicates that classifier $f$ contains bias regarding $s$. 

\begin{figure}[h]
	\setlength{\abovecaptionskip}{2.0pt}
	\centering
	\includegraphics[width=0.4\columnwidth]{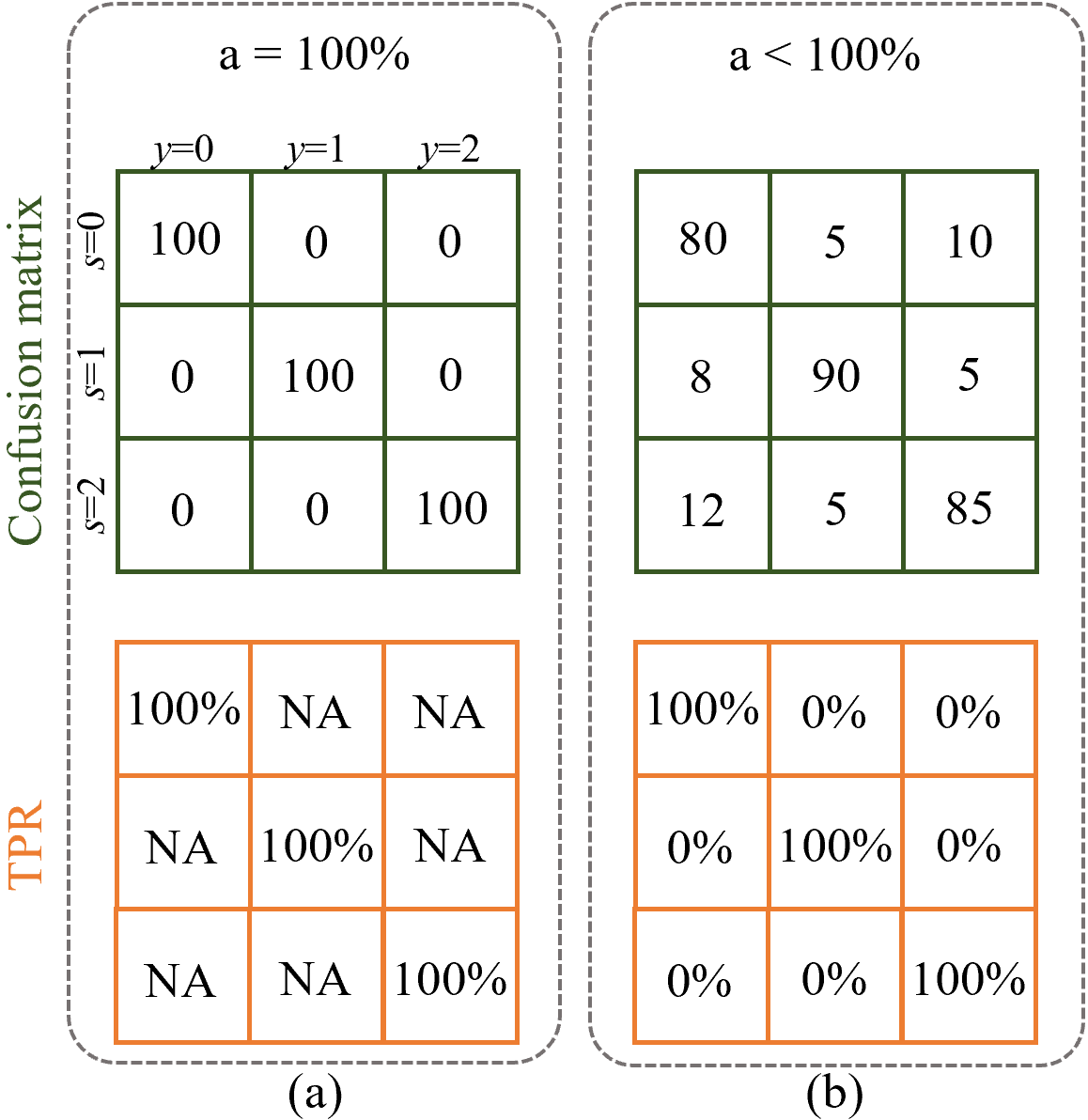}
	\caption{The hard confusion matrix and the TPRs. Take a 3-class classification task as an example, with each class contains 100 samples. 
	}\label{figapp:propostion}
\end{figure}

In our setting without bias annotations, we train an overfitted ERM and use its predictions $s$ as a general bias cues. 
Specifically, given a dataset $\Dc$ with spurious correlations, where minority groups are non-empty and target labels are correct, we train an ERM model $f$ on a small random subset of $\Dc$ and evaluate it on the full dataset, obtaining accuracy $a$.
\begin{itemize}
	\item If $a=100\%$, TPRs for each $(s,y)$ pair resemble Figure \ref{figapp:propostion}(a). This implies that bias cues are not preferred and $f$ likely relies exclusively on core features. No debiasing is needed.
	\item If $a<100\%$, overfitting occurs, though to different degrees. The TPRs within each class show severe violations of the EOF criterion (\eg Figure \ref{figapp:propostion}(b) for class $k=0$, $\text{Pr}\{\hat{y}=0|s=0,y=0\}\gg\text{Pr}\{\hat{y}=0|s\neq 0,y=0\}$), indicating that $f$ indeed captures and relies on the bias encoded in $s$.
\end{itemize}
Thus, in principle, as long as $a<100\%$, our method leveraging ERM overfitting reliably captures harmful implicit bias cues. 
Unlike \cite{li2022discover}, our approach directly identifies cues that maximally violate EOF without requiring interleaving optimization.

\textbf{R3W2: The hyperparameter choices in MST}    
  
In fact, we have conducted ablation studies on $\gamma$ in Figure 3(right) and discussed it in Section 4.4, which may have been overlooked.
To further validate its robustness across datasets and bias strengths, we include additional results on UrbanCars.
These results consistently show that $\gamma=10\%$ serves as a sweet spot for maximizing the smallest-mode recall.
Please refer to R1Q1 for our discussion regarding the use of the top $50\%$ high-confidence samples.

\textbf{R3W3: On MST’s ability to capture complex biases in multi-shortcut scenarios}

As we have pointed out in R3W1, we focus only on biases that are harmful --- i.e., those that cause ERM models to overfit and make incorrect predictions --- and our goal is to correct them. 
If the model overfits to “noise or irrelevant features” rather than physically interpretable biases, we treat such noise or irrelevant features as harmful bias and aim to balance them to improve ERM performance.

As demonstrated in Line 156 of the main manuscript, our model captures spurious cues that lead to overfitting and, consequently, incorrect predictions. These cues may correspond to interpretable shortcuts, combinations of multiple shortcuts, or entangled, uninterpretable patterns. Therefore, when multiple competing biases exist, MST can reveal the full bias structure, representing multiple competing biases within a single bias cue.

We have conducted experiments in Section 4.1 (Table 2) to demonstrate the effectiveness of our method in complex multi-shortcut scenarios, which may have been overlooked. For example, in UrbanCars, there are two competing shortcuts (background and co-occurring objects) and our method exhibits substantially less bias towards any specific background, co-object, or their combination, even outperforming methods that rely on multiple shortcut annotations.

Additionally, we compare the Recall of bias-conflicting modes on UrbanCars obtained by XRM, JTT, and our MST in Table \ref{tabapp:urbancar_mst}. The results show that even under multi-shortcut conditions, our method successfully identifies bias-conflicting samples covering all minority groups, whereas XRM fails to capture group $(0,1,1)$, and JTT fails to capture groups $(0,1,0)$, $(0,1,1)$, and $(1,1,0)$.

\begin{table*}[t]
	\centering
	\caption{
		Recall of minority groups in UrbanCars predictions by MST, XRM, and JTT. Group $(e_1, e_2, e_3)$: $e_1=0/1$ indicates urban/country car, $e_2=0/1$ indicates urban/country object, and $e_3=0/1$ indicates urban/country background.
		\label{tabapp:urbancar_mst}
	}
	\renewcommand{\arraystretch}{1.1}
	\resizebox{0.70\hsize}{!}{
		\setlength{\tabcolsep}{6pt}
		\begin{tabular}{c|c|c|c|c|c|c}
			\hline
			& (0,0,1) & (0,1,0) & (0,1,1) & (1,0,0) & (1,0,1) & (1,1,0) \\
			\hline
			MST & \textbf{45.79\%} & \textbf{58.42\%} & \textbf{70.00\%} & \textbf{100.00\%} & \textbf{64.55\%}& \textbf{28.57\%} \\
			\hline 
			XRM & 41.05\% & 30.51\% & 0.00\% & 60.00\% &10.12\% & 14.06\% \\
			JTT & 0.53\% & 0.00\% & 0.00\% & 10.00\% & 0.53\% & 0.00\%\\
			\hline
		\end{tabular}
	}
	\vspace{-0.4 cm}
\end{table*}

\textbf{R3W4: Quantitative Comparison of MST with XRM and DebiAN}.

The comparison with XRM and DebiAN on final debiasing performance was already provided in Table 1 and Table 2 of the main manuscript (Section 4.1), which may have been overlooked.
Similarly, the comparison with XRM and JTT on bias capturing was already presented in Figure4(b) and discussed in section 4.3, which also may have been overlooked. 
Theoretically, XRM trains its biased model on a random half of the training data, which contains far more bias-conflicting samples than ours, resulting in lower precision and recall on the smallest-mode. A similar explanation accounts for JTT's poor recall. DebiAN uses an alternating training scheme, where the classifier gradually mitigates biases during the discovery phase, making it difficult for its discoverer to reliably predict biases; therefore, we do not include DebiAN in the bias-capturing comparison.

To provide a more comprehensive evaluation, we have additionally included the F1-score in Figure 4(b). Our method consistently achieves the highest F1-score.

\textbf{R3W5: F1-score to evaluate MST mode partitions}.

Please refer to R3W4 for quantitative evaluation of MST-generated mode partitions. 

The effect of the MST's prediction quality on the subsequent FG-CCDB was already provided in Figure1(c) and may have been overlooked. Please refer to R2W4 for a detailed discussion.

\textbf{R3Q1: on further subdivision within modes or continuous weights}.

We appreciate the reviewer’s insightful suggestion. While a mode may contain potential substructures, our assumption of intra-mode homogeneity is not a theoretical requirement but a practical approximation, motivated by the following:
($i$) the "mode" definition is conditioned on both the predicted bias and the label $(s,y)$.
The auxiliary bias model partitions data according to the most dominant spurious patterns revealed by ERM overfitting. This ensures that samples assigned to the same mode share the most influential bias cues, which is sufficient for effective reweighting. In practice, these dominant bias cues account for the majority of generalization errors, while finer-grained variations within a mode have only marginal influence.
($ii$) Empirically, uniform per-mode weighting is stable and effective. We experimented with an alternative design (i.e., entropy-based intra-mode splitting that divides each mode into high-entropy and low-entropy subsets) but found it introduced noise and degrades performance.
For example, on Waterbirds, WGA drops from $90.56\%$ to $89.90\%$, and on cCIFAR10 with an extremely small bias-conflicting portion (the smallest group contains only 19 samples), performance drops from $55.28\%$ to $50.18\%$.
This suggests that finer intra-mode partitioning requires additional sub-bias cues to correctly guide matching, which are unavailable under the current setting. 

FG-CCDB focuses on mode-level bias amplification guided by dominant shortcuts. Incorporating a more detailed internal structure is beyond the scope of this work. We therefore consider mode-level homogeneity a reasonable and empirically validated design trade-off, with finer-grained mode modeling left as future work.

\textbf{R3Q2: Performance curves over additional iterations to demonstrate MST convergence}.

\begin{figure}[t]
	\setlength{\abovecaptionskip}{2.0pt}
	\centering
	\includegraphics[width=0.7\columnwidth]{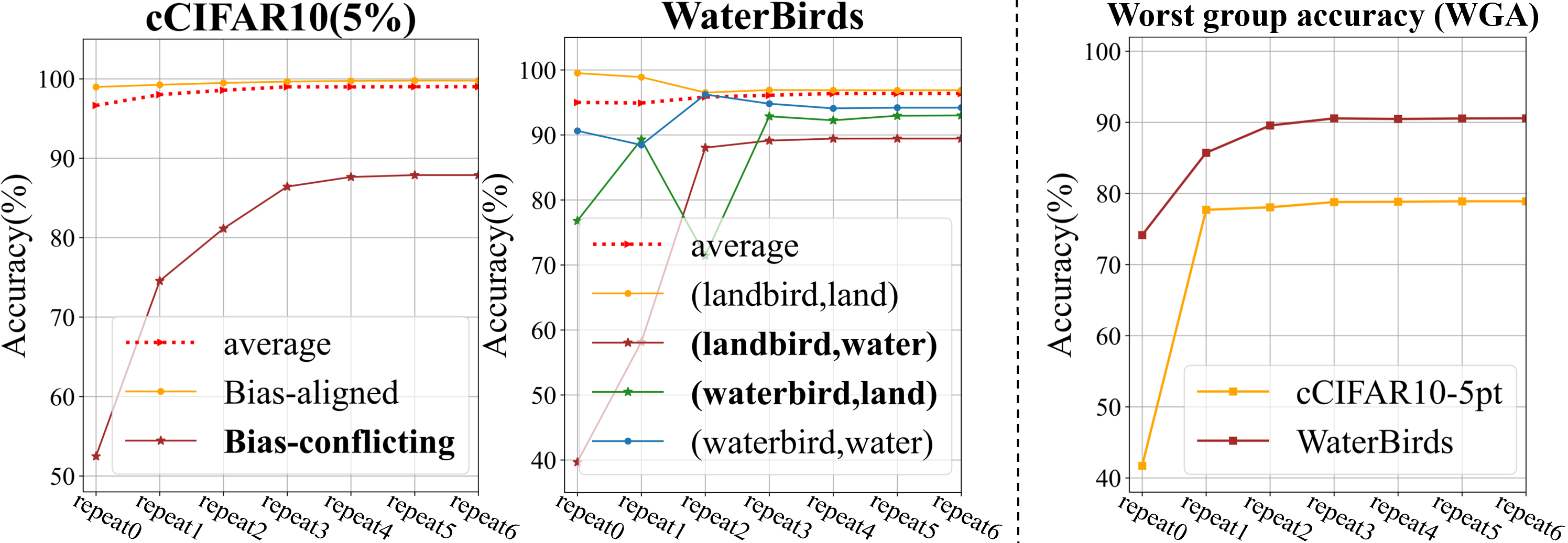}
	\caption{Mode prediction recall (left) and WGA under varying mode prediction quality (right) across repetitions of the ``bias enhancement learning'' procedure.. 
	}\label{figapp:repeating}
\end{figure}

We provide the mode partition and WGA results with additional repetition counts in Figure \ref{figapp:repeating}.
The results show that when the number of repetitions exceeds 3, the improvement in mode-partition accuracy slows down and eventually converges to a stable point. Correspondingly, the WGA remains nearly unchanged once the repetition count is greater than 2.

This behavior is expected. As repetitions progresses, bias-conflicting samples are gradually filtered out, causing the bias-aligned ratio of the selected training subset to increase and eventually stabilize. Once the learned bias model reaches a stable level of bias reliance, further top-confidence selection no longer changes the bias-aligned ratio, and the mode partition consequently remains unchanged.

\textbf{R3Q3: Computational cost of MST}.

To avoid misunderstanding, uLA is also a two stage method.
Compared to single-stage training methods like ERM, the additional training time mainly comes from MST. However, this overhead is acceptable in practical applications for the following reasons:
($i$) In each MST stage, we use only $10\%$, $50\%$, $50\%$, and $50\%$ of the training data, which significantly reduces the computational burden.
($ii$) We observe that the ERM model already exhibits strong bias reliance in the early training phase --- a phenomenon widely reported in prior works. Therefore, we set a small number of epochs for each MST stage.
The main computation cost are compared in Table \ref{tabapp:cost1}, 
and the running-time of MST is summarize in Table \ref{tabapp:cost2}.

\begin{table*}[t]
	\centering
	\begin{minipage}[t]{0.45\textwidth}
		\centering
		\caption{
			The computation cost of compared methods on cCIFAR10.
			\label{tabapp:cost1}
		}
		\renewcommand{\arraystretch}{1.2}
		\resizebox{0.98\hsize}{!}{
			\setlength{\tabcolsep}{1pt}
			\begin{tabular}{c|c|c|c}
				\hline
				& Our & ERM & uLA  \\
				\hline
				Bias discovery & 80 epochs & NA & 500 epoch \\
				\hline 
				Debiasing & 5000 iters$\approx$28 epochs & 300 epoch & 500 epochs\\
				\hline
			\end{tabular}
		}
	\end{minipage}
	\hfill
	\begin{minipage}[t]{0.54\textwidth}
		\centering
		\caption{
			The running time (hour) of MST, evaluated on a single NVIDIA A40 GPU.
			\label{tabapp:cost2}
		}
		\renewcommand{\arraystretch}{1.2}
		\resizebox{0.98\hsize}{!}{
			\setlength{\tabcolsep}{10pt}
			\begin{tabular}{c|c|c|c|c}
				\hline
				& cCIFAR10 & Waterbirds & CelebA & UrbanCars  \\
				\hline
				MST & 0.27h & 0.35h & 1.26h & 0.14h\\
				\hline
			\end{tabular}
		}
	\end{minipage}
	\vspace{-0.4 cm}
\end{table*}



\end{document}